\documentclass{article} 
\usepackage{iclr2025_conference,times}

\usepackage{amsmath,amsfonts,bm}

\def\eqref#1{equation~\ref{#1}}

\def\1{\bm{1}}

\DeclareMathAlphabet{\mathsfit}{\encodingdefault}{\sfdefault}{m}{sl}
\SetMathAlphabet{\mathsfit}{bold}{\encodingdefault}{\sfdefault}{bx}{n}

\usepackage[breaklinks=true]{hyperref}
\usepackage{url}
\usepackage{multirow}  
\usepackage{array}     
\usepackage{graphicx}  
\usepackage{pifont}

\usepackage{booktabs}
\usepackage{xspace}
\usepackage{tcolorbox}
\usepackage{wrapfig}

\tcbuselibrary{skins, breakable}

\title{AIGS: Generating Science from AI-Powered Automated Falsification}

\author{
Zijun Liu$^{1}$\thanks{indicates equal contribution.}$^{\phantom{*}}$, Kaiming Liu$^{1*}$, Yiqi Zhu$^{1*}$, Xuanyu Lei$^{1,2*}$, Zonghan Yang$^{1*}$,\\
$\phantom{,}$\textbf{Zhenhe Zhang}$^{1}$\textbf{,} \textbf{Peng Li}$^{2}$\textbf{,} \textbf{Yang Liu}$^{1,2}$\\
$^1$Department of Computer Science \& Technology, Tsinghua University\\
$^2$Institute for AI Industry Research (AIR), Tsinghua University\\
}

\newcommand{\AIGSFull}{AI-Generated Science\xspace}
\newcommand{\AIGS}{AIGS\xspace}
\newcommand{\AIGSSys}{\textsc{Baby-AIGS}\xspace}
\newcommand{\ProposalAgent}{\textsc{ProposalAgent}\xspace}
\newcommand{\ExperimentAgent}{\textsc{ExpAgent}\xspace}
\newcommand{\LiteratureAgent}{\textsc{LiteratureAgent}\xspace}
\newcommand{\ReviewAgent}{\textsc{ReviewAgent}\xspace}
\newcommand{\FalsificationAgent}{\textsc{FalsificationAgent}\xspace}
\newcommand{\DSL}{\textsc{DSL}\xspace}

\iclrfinalcopy 

\begin{document}

\maketitle

\vspace{-2pt}
\begin{abstract}

Rapid development of artificial intelligence has drastically accelerated the development of scientific discovery. Trained with large-scale observation data, deep neural networks extract the underlying patterns in an end-to-end manner and assist human researchers with highly-precised predictions in unseen scenarios. The recent rise of Large Language Models (LLMs) and the empowered autonomous agents enable scientists to gain help through interaction in different stages of their research, including but not limited to literature review, research ideation, idea implementation, and academic writing. However, AI researchers instantiated by foundation model empowered agents with full-process autonomy are still in their infancy. In this paper, we study \underline{\textbf{\textit{\AIGSFull}}} (\AIGS), where agents independently and autonomously complete the entire research process and discover scientific laws. By revisiting the definition of scientific research~\citep{Popper1935-POPTLO-7}, we argue that \textit{\textbf{falsification}} is the essence of both human research process and the design of an \AIGS system. Through the lens of \textit{falsification}, prior systems attempting towards \AIGSFull either lack the part in their design, or rely heavily on existing verification engines that narrow the use in specialized domains. In this work, we propose \AIGSSys as a baby-step demonstration of a full-process \AIGS system, which is a multi-agent system with agents in roles representing key research process. By introducing \FalsificationAgent, which identify and then verify possible scientific discoveries, we empower the system with explicit \textit{\textbf{falsification}}. Experiments on three tasks preliminarily show that \AIGSSys could produce meaningful scientific discoveries, though not on par with experienced human researchers. Finally, we discuss on the limitations of current \AIGSSys, actionable insights, and related ethical issues in detail.\footnote{Official Website: \url{https://agent-force.github.io/AIGS/}. Code is released at \url{https://github.com/AgentForceTeamOfficial/Baby-AIGS}.} 

\end{abstract}

\begin{figure}[htbp]
    \centering\includegraphics[width=0.95\linewidth]{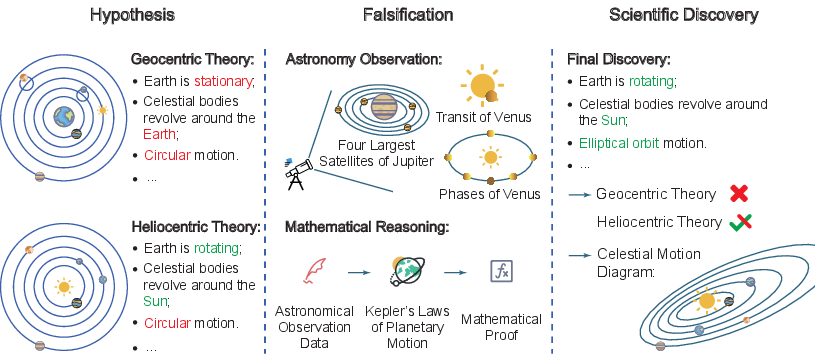}
    \vspace{-3pt}
    \caption{Examples of scientific research processes conducted by human researchers. Explicit falsification serves as a vital stage to falsify or verify the proposed hypotheses from either empirical or theoretical experiments, leading to the ultimate scientific discovery.} 
    \label{fig:CCF}
\end{figure}

\tableofcontents

\section{Introduction}
\label{sec:intro}

%

Deep learning has revolutionized scientific research~\citep{lecun2015deep,vaswani2017attention,jumper2021highly,achiam2023gpt}. Leveraging the enormous amount of experimental data, deep learning methods extract the underlying patterns in an end-to-end manner and effectively generalize to unobserved scenarios. The breakthroughs from deep learning in scientific domains, such as protein structure prediction~\citep{jumper2021highly}, gravitational wave detection~\citep{george2018deep}, and plasma control~\citep{degrave2022magnetic}, have received award-winning recognition. As a result, AI for Science has emerged as a highly-regarded research field~\citep{wang2023scientific}.

In the paradigm of AI for Science, AI primarily serves as a tool to assist researchers in making discoveries. With the rapid development of foundation models and autonomous agents~\citep{park2023generative}, AI techniques nowadays boast the capabilities of general-purposed textual understanding and autonomous interaction with the external world. These capabilities lead to the successful applications of AI-as-research-assistants, ranging from single-cell analysis~\citep{llmcell} to drug discovery~\citep{llm-drug}. The capability of providing research assistance leads to a more ambitious challenge: {\it Can foundation model-powered agents be autonomous researchers, independently completing the entire process of scientific discovery, thereby \textbf{transforming \underline{AI for Science} into \underline{\AIGSFull (\AIGS)}}?}

\looseness=-1 When constructing an \AIGS system with full-process autonomy, the desiderata of the system design should refer to the definition of the scientific research process itself. As stated by~\citet{Popper1935-POPTLO-7}, scientific research follows a systematic process of proposing novel hypotheses, conducting experiments through trial and error, and falsifying these hypotheses to conclude. While it is widely-believed that \textbf{\textit{creativity}} is indispensable in the process of research - which is also accounted in previous work~\citep{si2024can} - the central component of scientific research is \textbf{\textit{falsification}}: designing and executing experiments to validate or refute hypotheses, and falsified hypotheses pose positive contributions to scientific progress as well\footnote{\url{https://ml-retrospectives.github.io/}.}. Moreover, experienced researchers accumulate practical skills or reusable workflows~\citep{scientific-workflow} from hands-on experimentation, which eases the design and execution of experiments and hypothesis falsification. The abstraction of workflows in experiments enables effective reuse, which reflects a high level of \textbf{\textit{executability}} in scientific research. To recapitulate, a creative idea is the beginning of a piece of scientific research, which is followed by experiments and analyses to be conducted; executability forms the basis for falsification, and a sequence of logically consistent falsification processes turns a novel idea into scientific discoveries with genuine creativity. As a result, \textbf{\textit{falsification}} is the foundation of \AIGSFull, pillared by experimenting scaffolds accounting for \textbf{\textit{executability}} and targeting at the ultimate goal of research \textbf{\textit{creativity}}. 

\looseness=-1 Several preliminary works have been proposed to explore the potential of \AIGS, which can be roughly divided into three lines. In the first line, researchers evaluate and improve the capability of LLMs to generate research ideas with high \textbf{\textit{creativity}}~\citep{si2024can,hu2024nova}. The second line emphasizes the \textbf{\textit{executability}} of research experiments, e.g., benchmarks like MLAgentBench~\citep{liu2023ml} and MLE-Bench~\citep{chan2024mle} aim to evaluate the agentic ability of LLMs to achieve high performance on the provided benchmarks via code generation. These two lines of research investigate distinct sub-stages in the research process, failing to address the full-process autonomy. The third line of research attempts to construct end-to-end \AIGS systems that cover both \textbf{\textit{creativity}} and \textbf{\textit{executability}}. MLR-copilot~\citep{li2024mlrcopilotautonomousmachinelearning} takes existing research papers as input, and produces execution results by both generating ideas and implementing experiments. AI Scientist~\citep{lu2024aiscientist} further claims to be able to organize the generated ideas and experimental results into research papers as the output. This line of research arouses significant excitement in the community, but is feedbacked with controversy: Criticisms include the incremental nature of the generated knowledge ``tweaks", as well as the poor quality of the generated code and the paper presentation\footnote{\url{https://x.com/jimmykoppel/status/1828077203956850756}.}. Indeed, as further benchmarked by DiscoveryWorld~\citep{jansen2024discoveryworld}, DSBench~\citep{jing2024dsbenchfardatascience}, and ScienceAgentBench~\citep{chen2024scienceagentbenchrigorousassessmentlanguage}, an automatic \AIGS system that produces novel research in an end-to-end manner is still in the early stages, with significant gaps remains underexplored, especially in the area of autonomous \textbf{\textit{falsification}}. Furthermore, while specialized systems like AlphaGeometry~\citep{trinh2024solving} have achieved striking domain-specific performances, they rely heavily on the existing verification engines, which alleviate the need of autonomous falsification by AI itself.

In this work, we initiate \AIGSSys, our baby-step attempt toward a full-process \AIGS system. \AIGSSys comprises several LLM-powered agents, including \ProposalAgent, \ExperimentAgent, \ReviewAgent, \FalsificationAgent, etc., each responsible for distinct stages within the research workflow, mimicking the full-process human research that falsifies hypotheses based on empirical or theoretical results for scientific discoveries. \AIGSSys operates in two phases: the first phase iteratively refines proposed ideas and methods through enriched feedback, incorporating experimental outcomes, detailed reviews, and relevant literature. The second phase emphasizes explicit \textit{\textbf{falsification}}, a key feature absent in prior systems~\citep{lu2024aiscientist}, executed by \FalsificationAgent. Based on experimental results related to the proposed methodology, the agent identifies critical factors likely contributing to notable experimental phenomena, formulates hypotheses, and ultimately produces scientific insights verified through ablation experiments.
Additionally, we introduce a Domain-Specific Language (\DSL)~\citep{10.1145/1118890.1118892} for \ProposalAgent to articulate ideas and methodologies in an executable format, enhancing research \textit{\textbf{executability}}—particularly during experiments. We observe that multi-sampling proposals combined with re-ranking based on validation benchmarks can enhance the \textit{\textbf{creativity}} of methodologies developed during \AIGSSys’s first phase. We apply \AIGSSys across three tasks: data engineering, self-instruct alignment, and language modeling. Preliminary experimental results indicate that \AIGSSys can autonomously produce meaningful scientific discoveries from automated falsification, supported by qualitative analysis. We also observe consistent performance improvements during the iterative refinement process in \AIGSSys. Nevertheless, the performance of \AIGSSys still lags behind that of experienced researchers in top academic venues, suggesting avenues for further enhancement.

\section{The Development of AI-Accelerated Scientific Discovery}
\label{sec:literaturereview}

\begin{figure}[htbp]
    \centering
    \includegraphics[width=\linewidth]{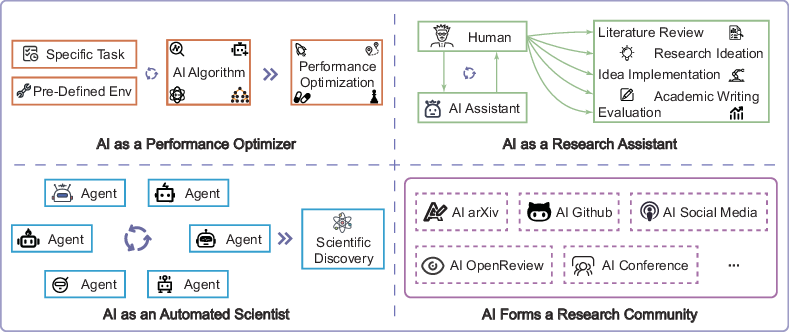}
    \caption{Overview of the four paradigms of AI-accelerate scientific discovery systems.}
    \label{fig:four-stages}
    \vspace{-1em}
\end{figure}

In this section, we review and envision the development of AI-accelerated scientific discovery as four paradigms (Figure~\ref{fig:four-stages}): (I) \textbf{AI as a Performance Optimizer}, where deep neural networks are trained with large-scale observation data in a specific scientific problem to extract the patterns in an end-to-end manner. In this paradigm, the AI techniques are used to optimize the specific prediction / regression performance in the pre-defined scientific problem with the consideration of out-of-domain generalization. (II) \textbf{AI as a Research Assistant}, where LLM-driven research copilots are used to assist the human research process. The synergy between Paradigm (I) and (II) forms the AI-powered acceleration of scientific discovery nowadays. (III) \textbf{AI as an Automated Scientist}. In this regime, foundation model empowered agents with scientist-like behavior should complete the entire research process, ranging from the initial idea proposal to the ultimate delivery of the scientific findings. (IV) \textbf{AI Forms a Research Community}. Upon the prosperity of fully-autonomous AI researchers depicted in the previous stage, we envision the collaborations among the agentic researchers foster an AI-formed research community. 

\subsection{AI as a Performance Optimizer: Discoveries in Specific Tasks}

With the rise of deep learning, AI has significantly impacted scientific discoveries across various fields, particularly in optimizing specific tasks by exploring well-defined search spaces or extracting patterns from piles of data. Utilizing specialized deep learning models, scientific breakthroughs continue to emerge across diverse fields, including accurate protein structure prediction~\citep{jumper2021highly, abramson2024accurate}, drug discovery and materials design~\citep{gilmer2017neural, juan2021accelerating}, and the simulation of physical systems~\citep{sanchez2020learning}. Moreover, a longstanding open problem in mathematics has been resolved through training a specialized Transformer-based expert~\citep{alfarano2024global}. It is widely recognized that deep learning models are highly effective in learning representations and patterns from data, enabling scientific discovery when appropriately guided.

Large Language Models (LLMs), equipped with extensive world knowledge and advanced reasoning, are emerging as increasingly creative and autonomous agents. They have demonstrated remarkable proficiency in autonomously developing evolutionary strategies for instruction datasets~\citep{zeng2024automaticevol}, identifying and rectifying their own weaknesses~\citep{cheng2024autodetect, mcaleese2024llmbugs}, and optimizing organizational structures for improved efficiency~\citep{zhang2024aflow, hu2024automatedagents}, highlighting their potential for performance optimization through structured search. Beyond language tasks, their creativity contributes to impressive discoveries in scientific fields. Via scientifically oriented, logically organized searches, LLMs can be guided to discover mathematical solutions~\citep{romera2024mathematical} and physical equations~\citep{ma2024llmphysics, shojaee2024llmsr}. Augmented with specialized tools and verification engine, LLMs are capable of solving advanced geometry problems~\citep{trinh2024solving}, designing chemical reactions~\citep{chen2024chemistxlargelanguagemodelempowered} and discovering novel materials~\citep{bran2023chemcrow, ghafarollahi2024protagents}.

\subsection{AI as a Research Assistant: Co-pilot in Human-AI Collaboration}

Equipped with expanding scientific knowledge and generative capabilities, LLMs gradually exhibit great potential to assist researchers at various stages of the research process. 

\textbf{Literature review} is a fundamental but tedious step for scientific research, highlighting the need for autonomous agents for this task. Advanced LLMs are employed to identify relevant literature for a given research topic and generate structured summaries~\citep{haman2024using, huang2023role}. For instance, \citet{agarwal2024litllm} introduces a retrieval-augmented framework to produce reliable summaries based on latest studies. Furthermore, \citet{hsu2024chime} utilizes LLMs to organize scientific studies within hierarchical structures and \citet{li2024chatcite} develops an agentic pipeline that produces comparative literature summaries guided by human workflows. In summary, LLM-based agents have demonstrated the capability to produce readable and detailed literature reviews. 

For \textbf{research ideation}, LLMs are employed to generate reasonable hypotheses~\citep{wang2023hypothesis, qi2023large, zhou2024hypothesis} based on internal knowledge and supplementary inputs. To compare the quality of LLM-generated ideas with human experts, a large-scale human study~\citep{si2024can} finds that LLMs can generate research ideas of higher novelty but slightly weaker feasibility. Furthermore, \citet{kumar2024can} and \citet{girotra2023ideas} evaluate the idea generation capabilities of different LLMs and recognize their potential to serve as the sources of inspiration. To enhance LLM-driven ideation, \citet{baek2024researchagentiterativeresearchidea}, \citet{nigam2024acceleron} and \citet{nigam2024interactive} develop multi-agent ideation frameworks based on scientific literature, generating novel research proposals to accelerate the life-cycle of research process. Despite these advancements, generating ideas that balance both novelty and feasibility remains a significant challenge for LLM-based agents~\citep{si2024can}. To evolve initial proposals into validated knowledge therefore demands substantial effort.

The attempts in AI-assisted \textbf{idea implementation and auto-experimentation} are usually conducted as repo-level coding tasks, given the growing coding capabilities of LLMs. Focused on research-related repo-level coding, \citet{jimenez2024swebench}, \citet{liu2023ml} and \citet{chan2024mle} present challenging coding benchmarks targeting machine learning and software engineering tasks. Meanwhile, \citet{yang2024swe}, \citet{wang2024opendevin} and \cite{tao2024magis} leverage agentic collaboration to automated coding from language instructions, offering promising avenues to reduce researchers' coding workloads and enhance efficiency. However, the vision for agents to autonomously implement novel ideas and conduct experiments end-to-end imposes significantly higher demands on coding agents. Current challenges include a relatively low success rate~\citep{lu2024aiscientist} and frequent misalignment between proposed ideas and their coding implementations, highlighting the need for improvements in both execution reliability and alignment with research objectives.

\looseness=-1 In the realm of \textbf{academic writing}, LLMs can be utilized for drafting structured outlines, refining human-written texts and presenting research findings. Recent studies~\citep{liang2024mapping, geng2024chatgpt} have demonstrated a steady increase for LLM usage in scientific writing. This trend presents both opportunities and challenges for academia. When properly used, LLMs could improve research efficiency and presentation; But when misused, risks emerge as well in terms of research integrity. Therefore, effective oversight through detection strategies~\citep{liang2024monitoring, yang2023survey, ghosal2023towards} and watermarking techniques~\citep{kirchenbauer2023watermark, zhao2023protecting, zhang2024remark} is both beneficial and necessary.

Additionally, following LLM-as-judge methods~\citep{zheng2023judging}, LLM-based agents are employed for comprehensive \textbf{evaluation} on research outputs~\citep{lu2024aiscientist, li2024mlrcopilotautonomousmachinelearning}. Comparing model-generated reviews with expert evaluations, researchers have evaluated the capabilities of LLMs to provide insightful and high-quality reviews by constructing meticulously annotated datasets~\citep{du2024llms} or training preference models~\citep{tyser2024ai}.
With multi-agent collaboration to promote in-depth analysis and constructive feedback, \citet{d2024marg}, \citet{jin2024agentreview} and \citet{yu2024automated} develop LLM-powered agent pipelines to perform paper reviews, helping researchers improve the quality of their papers. Furthermore, \citet{sun2024reviewflow} introduces a reviewing tool designed to support reviewers with knowledge-intensive annotations. In a notable development, ICLR conference adopt reviewer agents to provide constructive feedback on human-submitted reviews, showcasing a promising application of AI-assisted reviewing~\footnote{\url{https://blog.iclr.cc/2024/10/09/iclr2025-assisting-reviewers}.}. Recently, researchers also constructed benchmarks for AI as a research assistant at more than one stages above~\citep{lou2024aaar10assessingaispotential}. Overall, it is promising for LLMs to assist researchers with reliable research feedback.

\subsection{AI as an Automated Scientist: Towards End-to-end Scientific Discovery}

Structured in well-organized agentic pipelines, LLMs are increasingly capable of tackling complex tasks collaboratively, with end-to-end scientific research being one of the most ambitious and challenging applications. For instance, \citet{lu2024aiscientist} develops an iterative multi-agent framework that supports the entire research process, from proposing novel ideas to presenting polished findings. Similarly, \citet{li2024mlrcopilotautonomousmachinelearning} introduces an automated research system for machine learning, and \citet{manning2024automated} employs LLMs to simulate scientists for social science research. Beyond research systems, \citet{jansen2024discoveryworld} proposes a simulation environment designed to challenge agents in automated scientific discovery. Despite these advancements, current end-to-end research systems still fall short of generating falsifiable scientific findings, constrained by the capabilities of both designed framework and foundation models. While previous research~\citep{lu2024aiscientist} has yielded well-formulated outcomes, the vision of automated science discovery still requires further efforts.

\subsection{AI forms a Research Community: Enable Academic Swarm Intelligence}

Throughout human history, scientific progress has been greatly driven by collaboration, connection, and discussion among scientists, highlighting the power of a vibrant research community. We propose that a research community of AI scientists could significantly accelerate the pace of automated scientific discovery. For agentic community construction, LLM-driven agents can be organized to generate believable, human-like behaviors~\citep{park2022social, gao2024large, park2023generative} and to perform specific roles as assigned~\citep{li2024agent, hua2023war, xu2023exploring}. Although agent-based simulations of research communities are in an early developmental stage, they represent a promising avenue for the future of fully automated, AI-driven research.

\vspace{-5pt}
\section{\AIGSSys: A Baby Step Towards Full-Process \AIGS}
\label{sec:proofofconcept}

In this section, we elaborate how a baby-step system towards the full-process \AIGS is designed, in terms of design principles, overall system design, and detailed implementations.

\subsection{Design Principles of a Full-Process \AIGS System}\label{sec:principle}

The typical research process for human scientists~\citep{Popper1935-POPTLO-7} generally consists of two main stages: the pre-falsification stage, which encompasses exploration of research ideas, refinement of methodologies, and theoretical or empirical analysis, and the falsification stage, which involves hypothesizing scientific laws and validating these hypotheses based on theoretical or empirical findings. 
In research fields like machine learning, empirical results for falsification process, i.e. ablation studies, are collected after researchers design and build a system, and conduct experiments. In contrast, other fields operate differently. For example, in physics or biology, empirical results are gathered from instruments or equipment after the experimental design and execution, while in mathematics or the humanities, theoretical insights are often derived through logical reasoning or literature review rather than empirical experimentation. These root falsification processes of different subjects in distinct knowledge source. In this work, we primarily focus on empirical subjects that requires actual implementation of the methodology of a research idea to obtain empirical results for falsification process, e.g., machine learning, and leave other venues for future work. 

\looseness=-1 Human scientific research workflow above reflects the design principles of a full-process \AIGS system, which are \textit{\textbf{falsification}}, \textit{\textbf{creativity}}, and \textit{\textbf{executability}}. 
Each of the principle could be bridged with a specific stage in the research workflow: (1) Ablation studies are fundamentally established upon \textit{\textbf{falsification}}, verifying any key factors that contribute to significant experimental results. (2) To achieve smooth and consistent experimentation, we emphasize the importance of \textit{\textbf{executability}} of the proposed methodology, which serves as the basis for collecting empirical results for both method refinement and ablation studies. (3) \textit{\textbf{Creativity}} of the proposed idea is the overall objective of the research process, which could be achieved through idea refinement and be identified by \textit{\textbf{falsification}} process. 
\textbf{We especially argue that the process of \textit{falsification} is equally, if not more, critical in AI-powered automated scientific discovery systems}, given that human trust in AI-generated findings relies heavily on a convincing falsification process that ensures scientific rigor and transparency.


In sum, \textbf{\textit{falsification}} is the foundation of a full-process \AIGS system, pillared by experimenting scaffolds accounting for \textbf{\textit{executability}} and targeting at the ultimate goal of high research \textbf{\textit{creativity}}. 

\subsection{\AIGSSys System Design}\label{sec:overall-system}

\begin{figure}[t]
    \centering
    \includegraphics[width=\linewidth]{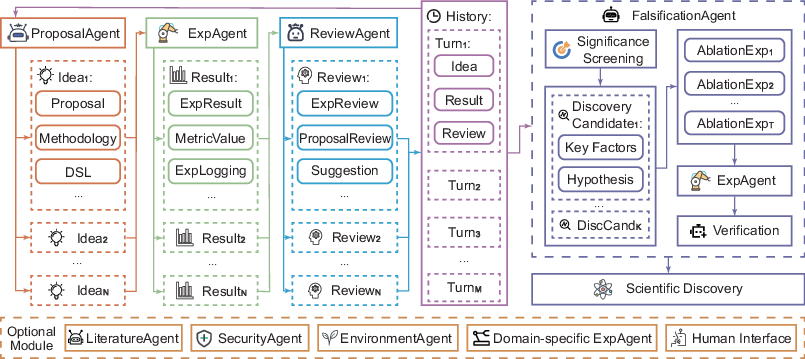}
    \caption{Overview of our \AIGSSys system design. The left part denotes \textbf{Pre-Falsification} phase, where \ProposalAgent iteratively refine the proposed idea and methodology based on empirical and verbose feedback from \ExperimentAgent, \ReviewAgent, etc. The iterative process summons multi-turn logs as the history context, based on which \FalsificationAgent could produce scientific discovery in the \textbf{Falsification} phase, as shown in the right part. Other modules are optional for the automated full-process research.}
    \label{fig:pipeline}
    \vspace{-1em}
\end{figure}




Heading towards a full-process system for automated scientific discovery, we present the design of \AIGSSys system in this section. We imitate the practice of human researchers and shape it into an LLM-powered multi-agent system. And we also take into account the capacity and behaviors of current foundation models to ensure the executability in implementation. 

The overall input for the system would be the topic of the research field, an accessible and configurable experiment environment, and other optional resources like a literature base; and the final outcome would be a verbal scientific discovery and the falsification process that support or falsify it. 
Following the principles in Section~\ref{sec:principle}, the \AIGSSys system operates in two phases (Figure~\ref{fig:pipeline}): 
\begin{enumerate}
    \item \textbf{Pre-Falsification}: This phase contains several stages, such as \textit{idea formation}, \textit{methodology design}, \textit{experiment execution}, \textit{result analysis}, etc., and operates iteratively for $M$ turns, aiming to explore and refine the proposed idea and method through feedback including experimental outcomes, reviews, etc. Specifically, the experimental results of turn 0 is from a trivial methodology at the default setting, e.g., no operation, identical mapping, etc. The multi-turn log of agent communications is recorded for \textbf{Falsification}. For better efficiency, this phase could be conducted with multi-sampling strategies, and the best ones for the next phase could be identified with experimental results. 
    \item \textbf{Falsification}: This phase aims to explicitly execute falsification by automating \textit{ablation studies}. The agent hypothesizes on what key factors are and how they might related to significant experimental phenomenon, and the ones pass $T$ designed ablation experiments are verified as final scientific discoveries. This could be also be $K$-parallel. 
\end{enumerate}



In the following sections, we elaborate important components of our \AIGSSys system. Ahead of specific modules, we introduce the Domain-Specific Language (\DSL)~\citep{10.1145/1118890.1118892}. In an \AIGSSys system, the \DSL acts a critical role to ensure the automated pipeline is errorless. Specifically, the \DSL is a human-designed descriptive language which can help interpret the proposed idea and methodology into executable experimental instructions through a pre-defined action space. For instance, in a deep learning task, the \DSL can directly be the codes that arrange training schedule of a model; While in a chemistry experiment, the \DSL can be the interface with a certain instrument or material. Consequently, the \DSL bridges the gap between formulation of proposed idea and experimentation, aligning the \AIGSSys system to the executability principle.

Here, we briefly depict the modules that construct the pipeline of \AIGSSys: 


\begin{itemize}
    \item \ProposalAgent is the module to propose ideas and methods within our system. It takes the detailed description of the task, the record of past experiments, and the review generated by \ReviewAgent as input, and outputs a proposal containing the idea, verbal and \DSL-format methodology, and other necessary components to carry out the experiment for \ExperimentAgent. It could iteratively interact with \ExperimentAgent to refine its proposal in order that the experiment can be successfully completed based on its proposal.
    \item \looseness=-1 \ExperimentAgent is responsible for experiment execution in the \AIGSSys system. It receives the proposal from \ProposalAgent and interprets \DSL the components relevant to the experiment into executable code. After execution, it transmits the experimental result as well as the whole process of the experiment to \ReviewAgent for review and analysis. 
    \item \ReviewAgent reviews the proposed idea and method based on the empirical results. It takes the whole record of both the experiments and the proposals as inputs, and generates the multi-granular review content. The review is then returned to \ProposalAgent for the next iteration of refinement. Through this iterative process between agents above in the \textbf{Pre-Falsification} phase, creativity of the proposed idea evolves in tandem.
    \item \FalsificationAgent is responsible for doing the ablation studies and deriving scientific discoveries as the final outcome. \FalsificationAgent takes the multi-turn log of all other agents as input. It has access to the record of the whole process of \textbf{Pre-Falsification} phase, and hypothesize possible key factors influencing significant experimental phenomenon based on empirical results. Then, it designs and conducts ablation experiments for $T$ times to verify the hypothesis, leading to final scientific discoveries.
    \item Other optional modules include \textsc{LiteratureAgent}, \textsc{SecurityAgent}, \textsc{EnvironmentAgent}, \textsc{Domain-specific ExpAgent}, and \textsc{Human Interface}. \textsc{LiteratureAgent} is responsible for gathering and providing relevant literature to support all other agents. \textsc{SecurityAgent} ensures safe experiment execution by identifying and preventing actions that may pose potential hazards or infringe upon intellectual property rights. \textsc{EnvironmentAgent} creates simulated environments to facilitate the testing and refinement of ideas, enabling more controlled and accurate scientific discoveries. \textsc{Domain-specific ExpAgent} is a customizable agent tailored for specific fields. \textsc{Human Interface} allows different agents in the system to ask human researchers for help when necessary.
\end{itemize}

We also acknowledge that the implementation of \AIGSSys at the current stage has various limitations towards a general functionable full-process \AIGS system. In Section~\ref{sec:actionableinsights}, we outline these limitations and discuss actionable insights for future improvements. 

\subsection{Detailed Implementation}

In the following sections, we elaborate on the the detailed implementation of our \AIGS system through \DSL, multi-sampling strategy, and three main agents: \ProposalAgent, \ReviewAgent, and \FalsificationAgent. The rest of optional modules have been omitted for the sake of clarity. 
In order to aid in the elaboration of the following sections, we present the research topic of data engineering~\citep{deita,chen2024alpagasus,li-etal-2024-one,long-is-more}, which requires \AIGSSys to identify key distinguishing features of datasets, and filter and extract high-quality data subsets. 
Implementation details are elaborated in Appendix~\ref{app:sys-detail} and Appendix~\ref{app:prompt}.


\subsubsection{Domain-Specific Language (\DSL)}
\label{sec:dsl}

\begin{figure}[t]
    \centering
    \includegraphics[width=\linewidth]{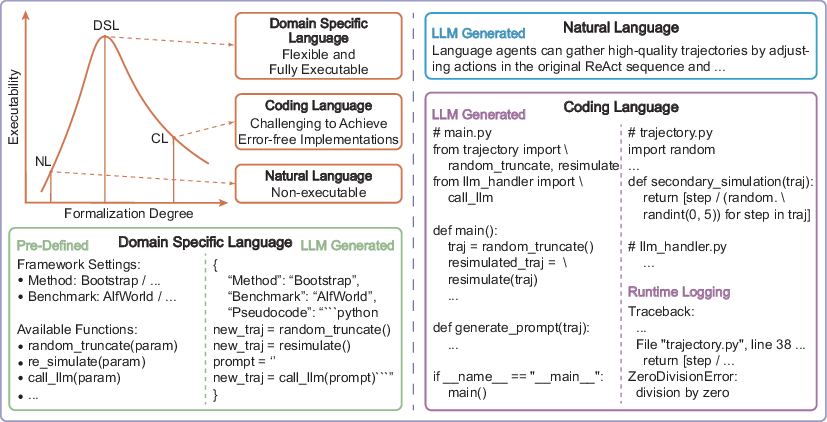}
    \caption{The relationship between formalization degree and system executability when expressing ideas through Natural Language~(NL), Coding Language~(CL), and Domain-Specific Language~(\DSL), illustrated with examples. NL expresses ideas in the simplest and most flexible form but is non-executable; CL offers greater precision but is challenging to achieve error-free implementation; \DSL achieves a better tradeoff between flexibility and executability.}
    \label{fig:idea-form}
    \vspace{-1em}
\end{figure}


\begin{wrapfigure}[27]{r}{0.465\textwidth}
    \vspace{-1.5em}
    \centering
    \includegraphics[width=0.98\linewidth]{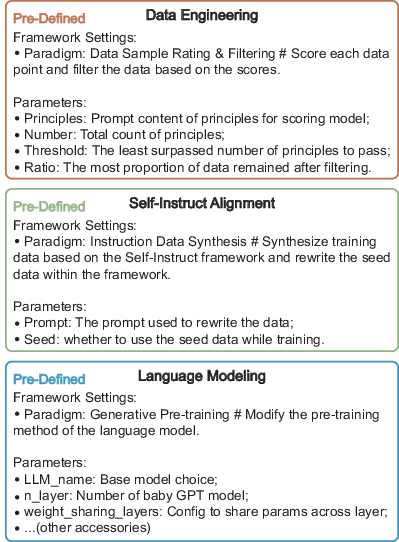} 
    \caption{The \DSL design in \AIGSSys for experimented research topics in Section~\ref{sec:exp-setup}. The full demonstration is in Appendix~\ref{app:exp-detail}.}
    \label{fig:dsl}
\end{wrapfigure}

A domain-specific language~\citep{10.1145/1118890.1118892} is created specifically for a particular application domain, providing greater expressiveness and ease of use within that domain compared to general-purpose languages, traditionally for programming languages. However, we observed that the situation is the same for agents in the \AIGS systems. 
When conducting scientific research, agents have access to a wide and diverse action space, making it challenging to perform error-free long-sequence actions for every stage of the research process, particularly when translating the methodology into executable actions for experimentation. For instance, in machine learning research, an agent may  edit multiple code files and manipulate large amount of data, as part of the methodology execution. However, limited by the current capacity of foundation models, it remains a severe challenge for agents to carry out the proposed experiment with both full-process autonomy and satisfiable success rates~\citep{jimenez2024swebench,chan2024mle, lu2024aiscientist} without dedicated interface design~\citep{yang2024swe,wang2024opendevin} or tool use~\citep{paranjape2023art,toolllm}.

In \AIGSSys, we extend the original definition of \DSL in programming to semi-structure objects with pre-defined grammars, making it a bridge that fills the gap between the proposed methodology and experimentation. The \DSL restricts the action space of the agents while maintaining the freedom for agents to conduct proposed methods at the same time, through dedicated design with human effort. To utilize the capabilities of current LLMs in natural language and function-level coding, we design the semi-structured grammar to be flexible between verbal instructions and structured statements. As shown in Figure~\ref{fig:idea-form}, the \DSL has both a higher degree of formalization and executability than natural language; compared to the coding language adopted in previous work~\citep{lu2024aiscientist}, though \DSL has a lower degree of formalization, with human effort, it exhibits higher executability and thus ensures successful execution of experiments, according to empirical analysis (Section~\ref{sec:exp-result}). However, when the grammar is poorly designed, the \DSL is likely to restrain the creativity of the system, because some ideas might not be able to be implemented, which is a limitation of \AIGSSys for future work. 

\looseness=-1 We present the pre-defined grammar of \DSL used in a few selected research topics in Figure~\ref{fig:dsl}. Under a specific paradigm related to the research topic, the grammar contains a series of parameters in either structured statement, e.g., code, integers, etc., or natural language, collectively depicting the methodology under the paradigm. \ProposalAgent would select a research paradigm when there are multiple, and fill out each parameter as required in the grammar. \ExperimentAgent is equipped with a pre-defined interpreter to translate the \DSL into executable code lines, or inputs to specific LLMs or other models. For instance, one parameter of the \DSL for data engineering is a few lines of data rating principles represented in natural language, and the model architecture parameters for language modeling still remains in codes, indicating the flexibility of \DSL design. Please refer to Section~\ref{sec:exp-setup} for detailed formulation of the research topics and topic-specific \DSL designs. 

\subsubsection{\ProposalAgent}
\label{sec:proposalagent}

\vspace{0.5em}

\begin{tcolorbox}[enhanced, breakable, title={An example of the proposal from \ProposalAgent}, colbacktitle=blue!55!black, coltitle=white, fonttitle=\bfseries, colframe=blue!55!black, colback=white, attach boxed title to top center={yshift=-2mm}, boxed title style = {sharp corners}]

    \begin{tcolorbox}[title={Idea \& Methodology}, colbacktitle=blue!50!white, coltitle=white, fonttitle=\bfseries, colback=blue!10!white, boxrule=0pt]
    \small
    \textbf{Idea}: ...Key issues identified include overly brief or excessively lengthy answers, lack of unique words, irrelevant content, poor adherence to instructions, lack of coherence, low keyword overlap, and poor sentiment balance...
    \tcblower
    \small
    \textbf{Methodology}: \textbf{Key metrics to observe} include the \textit{coherence of responses}, \textit{adherence to instructions}, \textit{relevance to the prompt}, \textit{depth of information provided}, \textit{clarity of instructions and responses}, \textit{engagement in the conversation}...
    \end{tcolorbox}
    \begin{tcolorbox}[title={Experiment Settings}, colbacktitle=blue!50!white, fonttitle=\bfseries, colback=blue!10!white, boxrule=0pt]
    \small
    Baseline: Iteration 0 \textit{(the trivial method)} \\
    Thought: ... we will filter the original dataset using the refined \DSL with weighted criteria. ... and this will help in identifying the initial impact of the new criteria on the raw data and ensure that the dataset is not overly biased by similarity...
    \end{tcolorbox}
    \begin{tcolorbox}[title={Hypothesis \& Related Feature}, colbacktitle=blue!50!white, fonttitle=\bfseries, colback=blue!10!white, boxrule=0pt]
    \small
    \textbf{Hypothesis}: After using the processed data, the model's performance on the MT-bench task will \textbf{improve significantly}. The model should produce longer, more detailed, and coherent responses, ... The responses should be rich in unique words, and demonstrate appropriate sentiment balance compared to the baseline.  
    \tcblower
    \small
    \textbf{Related Feature}: ... length of responses, keyword overlap, unique word count, and sentiment balance. 
    \end{tcolorbox}
    \begin{tcolorbox}[title={Rebuttal}, colbacktitle=blue!50!white, fonttitle=\bfseries, colback=blue!10!white, boxrule=0pt]
    \small
    The review should provide an overall view of the experiment result, focusing on whether the selected examples effectively demonstrate improvements in the key metrics. The review should compare the performance of the model before and after the data curation to highlight the impact of the methodology. Specific examples should be used to illustrate both improvements and remaining issues to provide ...
    \end{tcolorbox}
\end{tcolorbox}

\vspace{0.5em}


As the first step towards the scientific research, idea formation and methodology design usually lay the foundation for valuable insights or impactful discoveries from falsification process based on empirical results, i.e., \textbf{\textit{creativity}} in the \AIGS system. We refer to the corresponding module in \AIGSSys as \ProposalAgent, drawing inspiration from human practice of proposing an idea and formulating the methodology before starting the experiments.  

\ProposalAgent is important part of the pre-falsification phase. It takes the detailed description of research topic, the history log, including records of previous proposals and experiments, and the review from \ReviewAgent as the overall input, except for the first iteration, in which only the description of the research topic is the input to \ProposalAgent. 
As shown in the case above on the data engineering research topic, the output of \ProposalAgent includes 
\begin{itemize}
    \item the proposed \textit{Idea and Methodology}, that the former is a high-level thought and the latter is a semantically equal but concise description of instructions to be carried out in the experiment in natural language and \DSL format, aiming either to improve the experimental results or to advance towards scientific discoveries, 
    \item the configurable \textit{Experiment Settings}, such as specifying which previous iteration is considered the baseline for the current iteration, with other options specific to the experiment, 
    \item \textit{Hypothesis} on how would the experimental results change compared to and the most \textit{Related Feature} that may empirically reflect the hypothesis, which could guide \ReviewAgent to identify relevant components from all experimental results, 
    \item and \textit{Rebuttal} to the review from previous turns, except for the first iteration.
\end{itemize}
Thus, the formulation of \ProposalAgent could be expressed as:
\begin{equation}
\begin{aligned}
\textit{Proposal}^{(i)} &= \left\{\textit{Idea \& Method.}^{(i)}, \textit{Exp. Settings}^{(i)}, \textit{Hypo. \& Related Feat.}^{(i)}, \textit{Rebuttal}^{(i)}\right\}, \\
   & = \text{\ProposalAgent}\left(\textit{Research Topic} \mid \textit{History}^{(i)}\right), 1 \leq i \leq M, 
\end{aligned}
\end{equation}
where
\begin{equation}
\textit{History}^{(i)} = \begin{cases} 
\emptyset, & \text{if } i = 1 \\ 
\left\{\textit{Proposal}^{(j)}, \textit{Exp. Results}^{(j)}, \textit{Review}^{(j)} \right\}_{j=1}^{i-1}, & \text{if } 1 < i \leq M 
\end{cases},
\end{equation}
$i$ indicates the number of iteration, $M$ denotes the maximum iteration, $\ProposalAgent(\cdot \mid \cdot)$ indicates the agentic workflow, and \textit{Experimental Results} and \textit{Review} are from \ExperimentAgent and \ReviewAgent elaborated in Section~\ref{sec:reviewagent}. The \DSL format of the proposed methodology is illustrated in Appendix~\ref{app:exp-detail}.
Building upon the aforementioned components, \ProposalAgent puts forward a comprehensive yet highly executable proposal, which is then submitted to \ExperimentAgent for execution. Upon receiving the review form \ReviewAgent, \ProposalAgent can initiate the next iteration, either exploring a brand new direction or optimizing current experimental results. 

\subsubsection{\ReviewAgent}
\label{sec:reviewagent}

\begin{table}[t]
\resizebox{\linewidth}{!}{
\begin{tabular}{lp{1.2cm}p{8cm}l}
\toprule
\textbf{Metric} & \textbf{Level} & \textbf{Description} & \textbf{Execution} \\
\midrule
Length & \multirow{3}{*}{Corpus} & The length and word count of responses & \multirow{2}{*}{Pre-defined statistic function} \\
Keyword Overlap & & The keyword overlap between instructions and responses & \\ 
Sentiment & & The contained sentiment in model-generated responses & NLTK~\citep{bird-loper-2004-nltk} \\
\midrule
Worst Data Points & \multirow{2}{*}{Sample} & The worst rating samples compared with baselines & \multirow{2}{*}{Ranking \& reciting function} \\
Best Data Points & & The best rating samples compared with baselines & \\
\midrule
...... & Corpus / Sample & Other useful metrics generated by \ReviewAgent or pre-defined by researchers & Free-form code segment \\
\bottomrule
\end{tabular}
}
\caption{Examples of multi-level metrics for \ReviewAgent to empirically review the experimental results and the proposal from \ProposalAgent in the data engineering research.}
\label{tab:features_intro}
\end{table}


Drawing inspiration from human practice, we recognize that significant insights and breakthroughs often emerge from in-depth analysis of experiments and reflection on methodology based on empirical results. To facilitate this process, we design \ReviewAgent to analyze the experimental results and provide feedback to \ProposalAgent, iteratively improving the overall proposal.

In order to conduct a comprehensive and constructive review, \ReviewAgent performs analysis at different levels of granularity. For fine-grained analysis, \ReviewAgent examines comprehensive experimental logs, analyzing intermediate results from multi-level metrics which could be pre-defined by human researchers, e.g. performance indicators of the benchmark, or self-generated in code segment (examples for data engineering shown in Table~\ref{tab:features_intro}). The \textit{Review of the Experimental Results} identifies hidden patterns in the empirical details, resulting in fruitful low-level feedback mainly on experiment design and adjustment on the expectation of \ProposalAgent for the experimental results. For coarse-grained analysis, it evaluates the general validity and reasonableness of the methodology and hypothesis, providing \textit{Review of the} whole \textit{Proposal}. This review content serves as high-level advice on the idea and methodology, with the aim of provoking \ProposalAgent toward higher creativity. An example of a review of data engineering research is as follows: 

\vspace{0.5em}

\begin{tcolorbox}[enhanced, breakable, title={An example of the review from \ReviewAgent}, colbacktitle=blue!55!black, coltitle=white, fonttitle=\bfseries, colframe=blue!55!black, colback=white, attach boxed title to top center={yshift=-2mm}, boxed title style = {sharp corners}]
    \begin{tcolorbox}[title={Review of the Experimental Results}, colbacktitle=blue!50!white, fonttitle=\bfseries, colback=blue!10!white, boxrule=0pt]
    \small
    \textit{Summary and Actionable Insights}: Based on the comprehensive analysis of various features influencing the scores of responses in the Alpaca-GPT4 Database, here are the key findings and recommendations for optimizing the dataset... \\
    \textit{Key Insights}: \\
    1. Length and Word Count: High-quality responses tend to be longer, with word counts above 1000 for answers and around 15-20 words for queries. \\
    2. Conciseness: While length...
    \end{tcolorbox}

    \begin{tcolorbox}[title={Review of the Proposal}, colbacktitle=blue!50!white, fonttitle=\bfseries, colback=blue!10!white, boxrule=0pt]
    \small
    \textit{Evaluation of Current Research Components}: \\
    Your proposal effectively identifies key issues within the Alpaca-GPT4 dataset, such as... Additionally, the need for specific, measurable criteria for evaluating data points to improve... \\
    \textit{Suggestions}: 1. Data Distribution Analysis: Perform a quantitative analysis to understand the prevalence and distribution of these issues within your dataset...
    \end{tcolorbox}
\end{tcolorbox}

\vspace{0.5em}

Formally, the outcome of \ReviewAgent could be expressed as:
\begin{equation}
\begin{aligned}
\textit{Review}^{(i)} &= \left\{\textit{Review of the Exp. Results}^{(i)}, \textit{Review of the Proposal}^{(i)}\right\}, \\
   & = \text{\ReviewAgent}\left(\textit{Research Topic} \mid \textit{Proposal}^{(i)}, \textit{Exp. Results}^{(i)}, \textit{History}^{(i)}\right), 1 \leq i \leq M, 
\end{aligned}
\end{equation}
where $\ReviewAgent(\cdot \mid \cdot, \cdot, \cdot)$ indicates the agentic workflow, and \textit{Experimental Results} contain the benchmark results and other metric values extracted from experiments. 
In addition, human scientists derive valuable insights not only from a literature review and reasoning, but also through empirical analysis and detailed inspection of the experimental phenomenon, especially for subjects relying largely on empirical studies. Compared to previous work~\citep{lu2024aiscientist, su2024two} that improve ideation creativity primarily based on literature, our system advances this approach by introducing multi-granular review of experimental results and processes. We argue \textbf{the groundtruth of scientific laws root and get reflected in experimental outcomes, which could serve as process supervision} in our iterative refinement of the proposal in the pre-falsification phase, and might contribute to the overall creativity of \AIGSSys. Please refer to Section~\ref{sec:exp-result} for empirical analysis.

\subsubsection{Multi-Sampling Strategy}

In this section, we formalize the multi-sampling strategy employed in the pre-falsification phase of \AIGSSys system. This strategy is designed for better efficiency and quality of iterative exploration by parallel executing \ProposalAgent, \ExperimentAgent, \ReviewAgent, etc. for multiple threads, combined with reranking to retain the most promising threads for further exploration. 

As shown in Figure~\ref{fig:pipeline}, the multi-sampling strategy operates orthogonal to the iterative refinement of the proposal, where the pre-falsification process of each iteration $i$ involves parallel sampling across $ N $ threads, and each sampled thread represents a full pre-falsification process, including ideation, experimentation, reviewing, etc. Formally, let $ \mathcal{S}^{(i)} = \{ s_1^{(i)}, s_2^{(i)}, \dots, s_N^{(i)} \}, i=1, ..., M $ represent the set of threads sampled in iteration $ i $. Each sample $ s_j^{(i)}, j=1, ..., N $ undergoes experiments and reranking based on pre-defined criteria, and only a subset with top-ranked samples $ \mathcal{S}^{(i)}_{\text{top}} \subset \mathcal{S}^{(i)} $ of size $ N_s $ is retained for the next iteration. The process can be summarized as follows: 

\begin{enumerate}
    \item \textbf{Sampling Step}: In each iteration $ i $, the system generates $ N $ samples $ \{ s_1^{(i)}, s_2^{(i)}, \dots, s_N^{(i)} \} $ in parallel. If the former samples $\mathcal{S}^{(i-1)}_{\text{top}}$ are available, i.e., it is not the first iteration, each $ s_j^{(i)}, j=1, ..., N$ is generated by taking into account the historical log from the $\left(j \lfloor \frac{N}{N_s} \rfloor + 1\right)$-th sample of the previous $\mathcal{S}^{(i-1)}_{\text{top}}$ threads, i.e. $s_{j \lfloor \frac{N}{N_s} \rfloor + 1}^{(i-1)}$. 
    \item \textbf{Reranking}: All samples are reranked on the basis of the benchmarking result during experimentation. For simplicity, we adopt the average performance score of all benchmarks. 
    \item \textbf{Selection for Next Iteration}: After step 2, the samples are reranked and the top $ N_s $ samples are selected to form the set $ \mathcal{S}^{(i)}_{\text{top}} $ for the next iteration.
\end{enumerate}








\looseness=-1 Within \AIGSSys, the multi-sampling strategy with reranking is applied primarily in the \textbf{Pre-Falsification} phase, facilitating an extensive yet efficient exploration of ideas, methods, and experimental configurations. By iteratively narrowing down to the top candidates, this strategy effectively focuses resources on promising pathways. 
In Section~\ref{sec:exp-abl}, we empirically demonstrate the multi-sampling strategy, coupled with reranking, is essential for guiding the iterative process in \AIGSSys towards scientifically significant discoveries in an effective and potentially scalable manner. 

\subsubsection{\FalsificationAgent}\label{sec:falsi-agent}

In the research process, there is usually a gap between the experimental results indicating improvement in performance and the final conclusions of the scientific findings, and human researchers usually perform ablation studies to verify the authenticity of scientific discoveries. We term progress like this \textbf{\textit{falsification}}, which is a critical step towards full-process automated scientific discoveries. 

\begin{wrapfigure}{r}{0.5\textwidth}
    \vspace{-1.5em}
    \centering
    \includegraphics[width=0.98\linewidth]{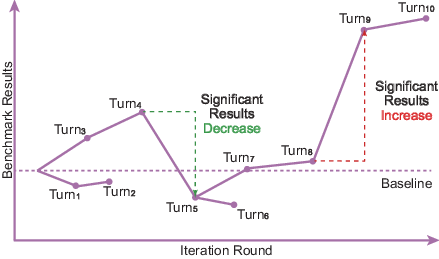} 
    \vspace{-0.5em}
    \caption{Illustration of ``Significance Screening'' on the history records of pre-falsification phase. The node of each turn represents the experimental results from the modified method. The ``Significance Screening'' process identifies results with significant performance increase or decrease for probably important scientific discoveries.}
    \label{fig:significance_screening}
\end{wrapfigure}

Recognizing the importance of \textbf{\textit{falsification}}, we introduce \FalsificationAgent, a novel component not present in previous work~\citep{lu2024aiscientist,su2024two}. \FalsificationAgent has access to all history records, including proposals from \ProposalAgent, experiment results from \ExperimentAgent, and reviews from \ReviewAgent. We hypothesize that important scientific discoveries are more likely to emerge from significant experimental phenomena, i.e. changes in results, thus, \FalsificationAgent in \AIGSSys first performs a ``Significance Screening'' to identify adjacent turns from pre-falsification phase with greatest performance discrepancies, as shown in Figure~\ref{fig:significance_screening}. Following this, \FalsificationAgent generates scientific discovery candidates from these selected turns. Then \FalsificationAgent generates the plans and the ablated methods for ablation experiments. We require that at most $T$ plans are made for each discovery candidate, indicating that at most $T$ ablation experiments will be conducted, and each ablation experiment focuses on the verification of a single factor that may influence the experimental result. Specifically, \FalsificationAgent must select an iteration from pre-falsification as the baseline for the ablation study, and \FalsificationAgent follows the ``Experiment Settings'' of the baseline, and modify the methodology according to the ablated factor. 
Attempting to reach a robust and reliable conclusion of the ablation study, both baseline and ablation experiments are repeated multiple times. \FalsificationAgent is given the complete record of these experiments to decide the validity of the associated scientific principle. If a particular discovery withstands this process and consistently produces results similar to those in the main experiment, it is regarded as a verified and valuable \textit{Scientific Discovery}. And it is falsified otherwise. 

Formally, the outcome of \FalsificationAgent, which is also the output of \AIGSSys, is:
\begin{equation}
\begin{aligned}
\textit{Scientific Discovery} & = \text{\FalsificationAgent}\left(\textit{Research Topic} \mid \textit{History}\right), 
\end{aligned}
\end{equation}
where
\begin{equation}
\textit{History} = \left\{\textit{Proposal}^{(i)}, \textit{Exp. Results}^{(i)}, \textit{Review}^{(i)} \right\}_{i=1}^{M},
\end{equation}
and $\FalsificationAgent(\cdot \mid \cdot)$ indicates the agentic workflow. 
We also provide an example on the data engineering research to better describe the different parts of the output of \FalsificationAgent in \AIGSSys as follows, in which specific parts of the methodology are ablated and reasonable conclusions are made based on the results of the ablation experiment:


\begin{tcolorbox}[enhanced, breakable, title={An example of the falsification process from \FalsificationAgent}, colbacktitle=red!55!black, coltitle=white, fonttitle=\bfseries, colframe=red!55!black, colback=white, attach boxed title to top center={yshift=-2mm}, boxed title style = {sharp corners}]
    \begin{tcolorbox}[title={Discovery Candidate}, colbacktitle=red!50!white, fonttitle=\bfseries, colback=red!10!white, boxrule=0pt]
    \textbf{Key Factor}: Importance of Context and Specificity.
    \end{tcolorbox}
    \begin{tcolorbox}[title={Ablation Experiment Plan}, colbacktitle=red!50!white, coltitle=white, fonttitle=\bfseries, colback=red!10!white, boxrule=0pt]
    \small
    Conduct an ablation study by systematically removing or altering one element related to context retention or specificity at a time. For example, test the impact of removing specific instructions or reducing context retention by limiting the number of conversational turns accessible to the model. This will help identify which specific factors within context and specificity contribute most significantly to model performance on MT-bench.
    \end{tcolorbox}
    \begin{tcolorbox}[title={Methodology}, colbacktitle=red!50!white, fonttitle=\bfseries, colback=red!10!white, boxrule=0pt]
    \small
    \textbf{Methodology for Ablation Experiments}:\{...``Principles'': ``...7. Responses should be concise and fall within the optimal length range (800-1500 characters).\textbackslash n8. Responses should engage the user naturally and be informative.\textbackslash n9. Weighting of each principle should be considered based on its importance to the downstream task.\textbackslash n10. Incorporate dynamic thresholding to adjust based on the number of data points passing the initial filter.''...\}
    \tcblower
    \small
    \textbf{Methodology for Baseline Experiments}:\{...``Principles'': ``...7. Responses should be concise and fall within the optimal length range (800-1500 characters).\textbackslash n\textcolor{red}{8. Responses should demonstrate context retention and follow multi-turn dialogue accurately.}\textbackslash n9. Responses should engage the user naturally and be informative.\textbackslash n10. Weighting of each principle should be considered based on its importance to the downstream task.\textbackslash n11. Incorporate dynamic thresholding to adjust based on the number of data points passing the initial filter.\textbackslash n\textcolor{red}{12. Break down complex criteria into more specific sub-criteria to capture nuances better. For example, 'context retention' can include sub-criteria like 'long-term memory' and 'contextual continuity.'\textbackslash n13. Evaluate the impact of each criterion through ablation studies.}''...\} \\ \\
    \textit{(\textbf{Highlights}: The parts related to the candidate scientific factor are ablated from the baseline methodology (marked in \textcolor{red}{red}) to perform ablation experiments for verification.)}
    \end{tcolorbox}
    \begin{tcolorbox}[title={Experiment Result}, colbacktitle=red!50!white, fonttitle=\bfseries, colback=red!10!white, boxrule=0pt]
    \begin{minipage}{\linewidth} 
    \small
    \centering
    \begin{tabular}{|l|c|c|c|c|}
    \hline
    \multirow{2}{*}{\textbf{Metric}} & \multicolumn{2}{c|}{\textbf{Ablation}} & \multicolumn{2}{c|}{\textbf{Baseline}} \\
    \cline{2-5}
    & \textbf{Trial 1} & \textbf{Trial 2} & \textbf{Trial 1} & \textbf{Trial 2} \\
    \hline
    Vicuna-Bench (Validation) ↑ & 7.1625 & 6.7500 & 6.475 & 6.5375 \\
    \hline
    MT-Bench (Test) ↑ & 4.10625 & 4.1125 & 4.05625 & 3.96875 \\
    \hline
    \end{tabular}
\end{minipage}
    \end{tcolorbox}
    \begin{tcolorbox}[title={Verification \& Scientific Discovery}, colbacktitle=red!50!white, fonttitle=\bfseries, colback=red!10!white, boxrule=0pt]
    \textbf{Verification}: The discovery regarding the importance of context and specificity in multi-turn conversations appears to have a negative influence on the quality of the data for this specific task. The ablation study results suggest that \textbf{limiting the number of conversational turns and focusing on fewer criteria can lead to better scores, which implies that the discovery might not positively affect the data quality as initially thought}.
    \tcblower
    \textbf{Scientific Discovery}: The true scientific discovery here is that while maintaining context and specificity is generally important, overemphasizing these elements may not always lead to improved performance in all tasks. Simplifying the criteria and focusing on essential conversational elements without excessive context retention can lead to better quality results in the context of the MT-bench task.
    \end{tcolorbox}
\end{tcolorbox}

\vspace{0.5em}

To our knowledge, \FalsificationAgent is the first agent within AI-accelerated scientific discovery systems capable of autonomously completing the falsification process, by independently proposing scientific discovery candidates, designing and executing ablation experiments, and performing verification. For a detailed qualitative analysis, see Section~\ref{sec:exp-result}. 




\subsection{Automated Full-Process Research Experiment}
\label{sec:exp-setup}

\subsubsection{Selected Research Topics}

We conduct experiments on three primary research topics in machine learning to evaluate \AIGSSys in autonomous full-process research. Formally, let $ \mathcal{D}_k = \{ (x_i, y_i) \}_{i=1}^N $ denote the $k$-th benchmark of a given ML problem, where $ x_i $ represents input features and $ y_i $ represents the corresponding labels. The goal is building a system $ f: \mathcal{X} \rightarrow \mathcal{Y} $ that maximizes metric functions $ \mathcal{L}_k(f(x), y) $ over all benchmark $ \mathcal{D}_k $. We split benchmarks into validation and test ones, and only the former is available in the pre-falsification phase, avoiding wrong scientific discoveries from over-fit results.

\paragraph{Data Engineering} 
Data engineering is a critical research topic that focuses on the identification, extraction, and processing of relevant data features that significantly influence model performance. We formulate the research goal as follows: Given a data set $\mathcal{H}$ that contains instruction-response pairs, the goal is to identify the key distinguishing characteristics of $\mathcal{H}$, which in turn enables the system to filter and extract high-quality data subsets $\mathcal{H}' \subset \mathcal{H}$ for the development of LLMs. This process is crucial to improving the quality and relevance of data for a wide range of areas, ensuring downstream tasks, such as in-context learning~\citep{brown2020language} and Supervised Fine-Tuning (SFT) for LLM alignment~\citep{NEURIPS2022_b1efde53}, are more effective. Specifically, we leverage Alpaca-GPT4 dataset~\citep{peng2023instruction} as the dataset $\mathcal{H}$. We follow previous work~\citep{deita,chen2024alpagasus,li-etal-2024-one,long-is-more} in this field and let the \AIGS systems write principles for LLMs to rate data samples and extract the top rated ones as the refined dataset. Thus, for \AIGSSys, we input the description of the topic and design the main \DSL as a list of required principles for the evaluation of the data sample and a threshold indicating the least number of principles that a data sample in the refined dataset has to pass. 

\paragraph{Self-Instruct Alignment}
The self-instruct alignment~\citep{wang-etal-2023-self-instruct} is a well adopted data synthesis paradigm for LLM alignment. The objective of this research topic is to synthesize a set of SFT data with high quality and diversity for LLM alignment~\citep{NEURIPS2022_b1efde53} by rewriting a seed set of data, thereby enhancing the performance of the fine-tuned model on this dataset. In the research process, an \AIGS system is required to construct an optimal set of instructions from a seed instruction dataset, which are used to generate an instruction-response dataset from LLMs. This dataset is then leveraged to refine the alignment of an LLM via SFT. In the experiment, we rewrite the original seed instruction set, and use the same LLM in instruction synthesis and response generation for SFT data. Specifically, for \AIGSSys, the \DSL is designed as an option whether to use the seed instruction set, and a list of requirements for the given LLM to generate instructions. 

\paragraph{Language Modeling} 
Language modeling is a core research topic in natural language processing that aims to improve the ability of a model to understand and generate human language. Currently, the mainstream approach is generative pre-training~\citep{radford2018improving}, and the objective is to maximize the perplexity of the next token prediction, i.e. minimize the model perplexity. The \AIGS system seeks to explore different architectural and training schedule modifications to enhance quality of language model pre-trained on large corpora. We designed \DSL of the \AIGSSys system as a set of constrained configurations of model architecture and training hyper-parameters. 

Each of these research topics requires unique methodological innovations of an \AIGS system to foster high \textbf{\textit{creativity}}, \textbf{\textit{executability}}, and \textbf{\textit{falsification}} capabilities. We demonstrate the pre-defined grammars of \AIGSSys in Figure~\ref{fig:dsl}. Please refer to Appendix~\ref{app:exp-detail} for detailed settings. 

\begin{table}[t]
\centering
\begin{tabular}{lccccc}
\toprule
             \textbf{Metric}              & \textbf{AVG}  & \textbf{STD}  & \textbf{P-Value} & \textbf{MIN} & \textbf{MAX} \\ \midrule\midrule 
\textbf{Importance Score} ($0\sim2$)  &      &      &         &     &     \\
\AIGSSys (Ours)        & 1.80 & 0.41 & 0.02    & 0.00   & \textbf{2.00}   \\
Top Conference             & \textbf{2.00} & 0.00 & ---     & \textbf{2.00}   & \textbf{2.00}   \\ \midrule 
\textbf{Consistency Score} ($0\sim2$) &      &      &         &     &     \\
\AIGSSys (Ours)        & 1.00 & 0.86 & 0.00    & 0.00   & \textbf{2.00}   \\
Top Conference             & \textbf{2.00} & 0.00 & ---     & \textbf{2.00}   & \textbf{2.00}   \\ \midrule 
\textbf{Correctness Score} ($0\sim2$) &      &      &         &     &     \\
\AIGSSys (Ours)        & 0.95 & 0.94 & 0.00    & 0.00   & \textbf{2.00}   \\
Top Conference             & \textbf{2.00} & 0.00 & ---     & \textbf{2.00}   & \textbf{2.00}   \\ \midrule 
\textbf{Overall Score} ($0\sim2$)     &      &      &         &     &     \\
\AIGSSys (Ours)        & 1.25 & 0.47 & 0.00    & 0.67   & \textbf{2.00}   \\
Top Conference             & \textbf{2.00} & 0.00 & ---     & \textbf{2.00}   & \textbf{2.00}   \\ \bottomrule
\end{tabular}
\caption{Statistic results of human evaluation on the falsification process in our data engineering research experiments.}
\label{tab:human}
\end{table}

\subsubsection{Evaluation Settings}

We evaluate \AIGSSys based on three key principles central to \AIGS systems as proposed in Section~\ref{sec:principle}: \textbf{\textit{falsification}}, \textbf{\textit{creativity}}, and \textbf{\textit{executability}}. We introduce the AI Scientist~\citep{lu2024aiscientist} as the baseline of the automated research system, and also select published literature from top conference as the baseline of research from experienced human researchers. 


\paragraph{Falsification} We assess \AIGSSys’s ability to perform falsification through human evaluation, focusing on the falsification process carried out by \FalsificationAgent. This process involves hypothesizing potential influencing factors, identifying the key variables that may impact experimental results, designing and conducting ablation experiments, and ultimately validating the real factors contributing to the experimental significance. The human evaluation is carried out by volunteer researchers with experience in publishing at top-tier conferences. Evaluators assess the falsification process based on three key dimensions, each scored on a scale from 0 to 2, with a higher score indicating better performance:
\begin{itemize}
    \item \textit{\textbf{Importance Score}}: This score reflects the importance of the scientific discovery candidate. It evaluates the extent to which the identified factors can influence the experimental results, considering their relevance and potential impact with the primary experiments.
    \item \textit{\textbf{Consistency Score}}: This score assesses whether the proposed ablation experiment plan is aligned with the identified scientific discovery candidate. It considers whether the experiments are designed to ablate the factor of interest and appropriately test the hypothesis.
    \item \textit{\textbf{Correctness Score}}: This score evaluates the accuracy of the final scientific discovery derived from the ablation studies. It considers whether the conclusions drawn from the results from ablation experiments are correct, based on the observed empirical results. 
    \item \textit{\textbf{Overall Score}}: This score is the average of all other dimensions mentioned above for each sample, serving as a comprehensive indicator of the quality of the falsification process. 
\end{itemize}
Additionally, several studies from the top conferences~\citep{deita, chen2024alpagasus, li-etal-2024-one, long-is-more} are included in the evaluation set to serve as a baseline. We conduct the evaluation on the data engineering research experiment, with statistic results shown in Table~\ref{tab:human}, where the p-values are obtained from a left-tailed Welch’s t-test on 20 samples against the top conference baseline and the gap is considered significant when $p<0.05$. Please refer to Appendix~\ref{sec:human_evaluator} for the experiment template provided to the evaluators in the human evaluation.

\paragraph{Creativity} We measure the creativity of \AIGSSys by evaluating the performance improvement of the proposed idea and methodology against the baseline result, i.e., the result from the trivial methodology on the test benchmarks. Here are the benchmark settings for each research experiment:
\begin{itemize}
    \item \textbf{Data Engineering}: For the refined dataset, we conduct 15-shot In-Context Learning (ICL)~\citep{jiang2024manyshot} and SFT for LLM alignment to evaluate the overall quality. We evaluate the ICL-aligned LLM on the Vicuna-Bench, as an efficient validation benchmark, and ICL- and the SFT-aligned LLM on the MT-Bench~\citep{zheng2023judging}, which are used as test benchmarks. The baseline of turn 0 uses the original Alpaca-GPT4 dataset~\citep{peng2023instruction}. We replicate AI Scientist with the same experiment template. Moreover, we replicate Deita~\citep{deita} as the human research of the topic from the top conference.  
    \item \textbf{Self-Instruct Alignment}: We also assess the aligned LLM on the Vicuna-Bench, as the validation benchmark, and the MT-Bench, as the test benchmark. The baseline of turn 0 is the result of the original self-instruct method~\citep{wang-etal-2023-self-instruct}. 
    \item \textbf{Language Modeling}: We pre-train a mini-sized language model with the modified architecture based on the configured training schedule, on three different training sets~\citep{karpathy2015shakespeare, hutter2006enwik8, mahoney2011text8}. The validation and test benchmarks are the perplexity of LM on the split validation and test sets. With reference to \citet{lu2024aiscientist}, we adopt the default settings of the nanoGPT project\footnote{\url{https://github.com/karpathy/nanoGPT}.} as the baseline. 
\end{itemize}
Results on all test benchmarks are in Table~\ref{tab:creativity-data}, Table~\ref{tab:creativity-align}, and Table~\ref{tab:creativity-lm}, for each topic, respectively. 

\paragraph{Executability} We evaluate the \AIGSSys system's stability to execute research ideas errorlessly from ideation to implementation, measured by the success rate of obtaining meaningful experimental outcomes and scientific insights, termed as Experiment Success Rate (Exp.~SR) and Overall Success Rate (Overall SR), respectively. We report the overall results on all research experiments on the three topics. AI Scientist as the baseline method, are also evaluated executability on the selected tasks in their original implementation~\citep{lu2024aiscientist}. Results are shown in Table~\ref{tab:success_rate}.  

\subsection{Quantitative and Qualitative Analysis}\label{sec:exp-result}

\paragraph{\AIGSSys could produce valid scientific discoveries with falsification process.}

To validate the falsification process in \AIGSSys, we assess its ability to perform ablation studies and identify causative factors for experimental results. The qualitative analysis in Table~\ref{tab:human} shows that \FalsificationAgent could produce valid scientific discoveries in current design, as the maximum value of each metric is tied to the top-conference baseline, contributing positively to the automation of scientific insights. However, there are two critical findings that indicate further improvement is needed. (1) The average value of the importance score is higher than the consistency and correctness score, indicating that \FalsificationAgent could identify important factors potentially related to a scientific discovery but failed to design a concrete experiment plan and verify the hypothesis. The failure could be attribute to the capacity of foundation model or the lack of high-quality demonstration of experiment design in prompts. (2) The p-values indicate that the falsification process of \AIGSSys is significantly less satisfactory than the existing literature from top conferences from human perspectives, which emphasizes the importance of designing user-friendly interfaces besides refining the design of ablation experiments. Also, we acknowledge that the scale of the study is small compared to \citet{si2024can}, which requires future effort. 

\begin{table}[t]
\centering
\begin{minipage}{0.42\linewidth}
\centering
\resizebox{\linewidth}{!}{
\begin{tabular}{lcc}
\toprule
\multirow{2}{*}{\textbf{Method}}  & \multicolumn{2}{c}{\textbf{MT-Bench} ↑}  \\\cmidrule{2-3} 
   &  15-shot ICL  &  SFT  \\ \midrule\midrule
   Baseline (Turn 0)  &  4.18   &  4.53 \\ 
   AI Scientist &  4.36  &  4.67 \\
   \textbf{\AIGSSys (Ours)}  &  \textbf{4.51}  & 4.77  \\ \midrule
   Top Conference &  4.45  &  \textbf{5.01} \\ \bottomrule
\end{tabular}}
\end{minipage}
\hfill
\begin{minipage}{0.56\linewidth}
\small
\begin{tcolorbox}[title={Methodology Summarization (Data Engineering)}, colbacktitle=blue!50!white, fonttitle=\bfseries, colback=blue!10!white, boxrule=0pt]
1. Rate the response based on its contextual coherence, ensuring it logically follows the conversation. \\
2. Evaluate the relevance by checking if the answer stays on-topic with minimal digression. \\
3. Check for logical reasoning in explanations, ensuring the response is not just factual but also thoughtful. \\
4. Consider if the complexity and detail match the question's requirements, avoiding oversimplification. \\
5. Finally, evaluate the tone for politeness, clarity, and natural conversational flow.
\end{tcolorbox}
\end{minipage}
\caption{Benchmarking results on the test benchmarks of the data engineering research experiment (left) and a summarization of the corresponding proposed methodology from \AIGSSys (right).}\label{tab:creativity-data}
\end{table}

\begin{table}[t]
\centering
\begin{minipage}{0.34\linewidth}
    \centering
\resizebox{\linewidth}{!}{
\begin{tabular}{lc}
\toprule
\textbf{Method}  &  \textbf{MT-Bench} ↑  \\ \midrule\midrule
   Baseline (Turn 0)  &   2.45  \\ 
   \textbf{\AIGSSys (Ours)}  & \textbf{3.26} \\ \bottomrule
\end{tabular}}
\end{minipage}
\hfill
\begin{minipage}{0.65\linewidth}
\small
\begin{tcolorbox}[title={Methodology Summarization (Self-Instruct Alignment)}, colbacktitle=blue!50!white, fonttitle=\bfseries, colback=blue!10!white, boxrule=0pt]
Make the instruction to cover different scenarios if it lacks specificity, clearer if ambiguous, aligned with natural conversations, and to contain a diverse range of task types if it lacks variety. 
\end{tcolorbox}
\end{minipage}
\caption{Benchmarking results on the test benchmark of the self-instruct alignment research experiment (left) and a summarization of the corresponding proposed methodology from \AIGSSys (right).}\label{tab:creativity-align}
\vspace{-1em}
\end{table}

\begin{table}[t]
\centering
\begin{minipage}{0.58\linewidth}
    \centering
\resizebox{\linewidth}{!}{
\begin{tabular}{lccc}
\toprule
\multirow{2}{*}{\textbf{Method}}  & \multicolumn{3}{c}{\textbf{Perplexity} ↓}  \\ \cmidrule{2-4}
 & shakespeare\_char &  enwik8 & text8  \\ \midrule\midrule
   Baseline (Turn 0)  &  \textbf{1.473}  &  1.003  & 0.974 \\ 
   \textbf{\AIGSSys (Ours)}  &   1.499 &  \textbf{0.984}  & \textbf{0.966} \\ \bottomrule
\end{tabular}}
\end{minipage}
\hfill
\begin{minipage}{0.41\linewidth}
\small
\begin{tcolorbox}[title={Methodology Summarization (Language Modeling)}, colbacktitle=blue!50!white, fonttitle=\bfseries, colback=blue!10!white, boxrule=0pt]
Reduce the dropout rate with more attention heads to increase model expressiveness.
And implement a cyclical learning rate and adjust the weight decay to regularize the model.
\end{tcolorbox}
\end{minipage}
\caption{Benchmarking results on the test benchmarks of the language modeling research experiment (left) and a summarization of the corresponding proposed methodology from \AIGSSys (right).}\label{tab:creativity-lm}
\end{table}

\paragraph{\AIGSSys demonstrates creativity during research idea exploration and refinement.}

Table~\ref{tab:creativity-data}, Table~\ref{tab:creativity-align}, and Table~\ref{tab:creativity-lm} show the results of the test benchmarks for \textit{data engineering}, \textit{self-instruct alignment}, and \textit{language modeling} research experiments, respectively, where \AIGSSys outperforms the baseline method, demonstrating the system's creativity in ideation and corresponding method design. For data engineering, \AIGSSys outperforms AI Scientist with a significant margin, demonstrating the effectiveness of the enriched feedback, including multi-granular metrics, verbose review on both experiment process and methodology design, etc., in exploring research idea. However, the result of SFT alignment is inferior than Deita~\citep{deita}, indicating that the lack of validation benchmarking of specific downstream tasks might result in an suboptimal outcome.

\begin{table}[t]
\centering
\resizebox{0.95\linewidth}{!}{
\begin{tabular}{ccc}
\toprule
\textbf{Method} & \textbf{Experiment Success Rate} (Exp.~SR) & \textbf{Overall Success Rate} (Overall SR) \\
\midrule\midrule
AI Scientist & 44.8\% & 29.2\% \\
\textbf{Baby-AIGS (Ours)} & \textbf{Almost 100\%} & \textbf{Almost 100\%} \\
\bottomrule
\end{tabular}}
\caption{Success rates on three selected tasks of AI Scientist and Baby-AIGS. Exp.~SR denotes the times a system successfully conducted experiments out of all trials, and Overall SR denotes the times a system produces the final scientific discoveries. Higher numbers indicate better executability.}
\label{tab:success_rate}
\vspace{-1em}
\end{table}

\paragraph{\AIGSSys has remarkable executability in experimentation and full research process.}

As shown in Table~\ref{tab:success_rate}, our quantitative analysis highlights significant improvements in executability, with \AIGSSys achieving nearly 100\% success rates in translating the generated ideas into experimental results and the final scientific discovery. This high executability, attributed to our \DSL design for errorless experimentation, prevents restarting from in-process failures and enables an efficient automated research process. Detailed API costs are elaborated in Appendix~\ref{app:token-anal}.

\subsection{Discussions}\label{sec:exp-abl}

\paragraph{\textbf{Q1: How do current LLMs perform in the falsification process?}}

Falsification~\citep{Popper1935-POPTLO-7} is essential in \AIGS systems as it provides a rigorous mechanism for verification of potential scientific discoveries, a core component in the scientific method. In \AIGSSys, \FalsificationAgent plays the corresponding role. Thus, it demands related abilities in the foundation model, such as reasonable hypothesis generation, ablation experiment design, summarization and self-correction based on input empirical results, etc. As shown in the case in Section~\ref{sec:falsi-agent} and Table~\ref{tab:human}, current LLMs are far from desired in the agentic workflow of \FalsificationAgent. Additionally, the constraints may come from the ability of the LLM to understand the environment outside \FalsificationAgent. For instance, from our observation, \FalsificationAgent seldom proposes experiment plans beyond the provided experiment templates. In this case, although \DSL makes sure the executability of the experimentation by omitting extra operations, the experiment process would differ from the original plan, thus creating inconsistency.

\paragraph{\textbf{Q2: Could \ReviewAgent serves as the \FalsificationAgent in the \AIGSSys system?}}

Previous work~\citep{lu2024aiscientist, su2024two, weng2024cycleresearcherimprovingautomatedresearch} typically involve an iterative process in research ideation and methodology refinement, along with designs similar to \ReviewAgent. This iterative exploration of research ideas and methods is indispensable. However, the review process of the changes in the methodology and the difference in the corresponding experimental results could not replace the explicit falsification process. In practice, we observed that behaviors of the exploration on the methodology of the AI Scientist and the pre-falsification phase of \AIGSSys are varied in a wide range, from a subtle adjustment of hyper-parameters to an abrupt rewriting of the whole idea. In quite a few cases, the changes of the experimental results resulting from the refinement of methodology could not represent clear single-factor patterns or scientific discoveries without dedicated ablation experiments, except for few instances. As a high-level explanation, we argue that \textit{an efficient and effective research process does not need to analyze the details of each possible change in methodology that has a random impact, but should analyze in detail those important changes that could possibly have a significant impact}. 

\begin{table}[t]
\centering
\small
\begin{tabular}{lcccccc}
\toprule
  \textbf{Method}     &   \textcolor{gray}{\textbf{Baseline}}     & \textbf{Turn 1}  & \textbf{Turn 2}  & \textbf{Turn 3} & \textbf{Turn 4} & \textbf{Turn 5} \\ \midrule\midrule 
 Multi-Sampling@1  & \textcolor{gray}{4.18}  &    3.68   &   4.01   &     4.05    &   3.88  &  3.90   \\ \midrule
 Multi-Sampling@32 &    \textcolor{gray}{4.18}  &  4.02	  &   4.05  &   4.50   &  4.51   & 4.42  \\ \bottomrule
\end{tabular}
\caption{Results on MT-Bench (15-shot ICL) of the ablation study on the multi-sampling strategy of our \AIGSSys system in the data engineering research experiment. $N$ in ``Multi-Sampling@N'' indicates the number of parallel threads of multi-sampling.}
\label{tab:abl-multi-sample}
\vspace{-1em}
\end{table}

\paragraph{\textbf{Q3: How does the \AIGSSys system boost creativity?}}

\AIGSSys enhances creativity by integrating a multi-sampling approach combined with re-ranking, allowing it to generate diverse research proposals and rank them based on validation benchmarks. We provide detailed results of an ablation study of this process in Table~\ref{tab:abl-multi-sample}. We observed that the performance on the test benchmark is steadily increasing with multi-sampling with large numbers of threads. 
This strategy is related to search-based inference-cost scaling methods~\citep{snell2024scalingllmtesttimecompute,brown2024largelanguagemonkeysscaling}. The insight is to pick random high-performing samples for better overall performance. 
However, since the objective of \AIGS is to discover science on a research topic, the reranking method here could be large-scale validation benchmarks indicating generalization performance, rather than reward-model-based~\citep{10.5555/3495724.3495977} or self-verification methods for a specific query. As depicted in Section~\ref{sec:reviewagent}, we argue that the groundtruth of scientific laws is rooted and reflected in benchmarking results from actual experiments, which could serve as process supervision, which could be more accurate than reward models. It explains how collapse in self-refinement-style methods~\citep{xu-etal-2024-pride} is avoided in this setting, which is also empirically validated through the ablation results. 

\paragraph{\textbf{Q4: Why could \DSL help with executability?}}

The use of a Domain-Specific Language (\DSL) in \AIGSSys facilitates executability by providing a structured and executable representation of ideas and methodologies proposed by \ProposalAgent. \DSL enhances the system’s ability to translate complex scientific workflows into actionable experiment plans. As shown in Table~\ref{tab:success_rate}, \DSL significantly improved success rates in generating scientific discoveries, regardless of correctness, underscoring its role in achieving high executability. We acknowledge that the design of \DSL requires human effort and might not be able to cover all possible method implementations. However, we believe it is a promising interface between agents and experimentation in full-process research.

\section{Limitations and Actionable Insights}
\label{sec:actionableinsights}

Envisioning the future of \AIGSFull systems powered by foundation models in real-world, in this section, we enumerate a few limitations for current \AIGSSys system and provide insights on the next steps of research for \textbf{\AIGS}.

\paragraph{Balance idea diversity and system executability.}
As discussed in Section~\ref{sec:dsl}, the design of the \DSL enhances the system executability but may constrain the idea diversity.
Achieving a balance between idea diversity and system executability requires further empirical analysis.
One potential avenue is enabling agents to develop their own DSLs, which could enhance the executability of generated ideas without diminishing their diverse potential.

\paragraph{Establish systematic mechanisms for evaluation and feedback.}
The quality of \AIGS system depends heavily on rigorous evaluation of prior proposals, methods, and results.
Current approaches often adopt a peer review format, leveraging LLMs to generate feedback on results and guide future optimization~\citep{lu2024aiscientist, yu2024automated, jin2024agentreview}.
However, it remains unclear whether this method is the most effective for large-scale research settings.
Future work should explore systematic mechanisms to analyze outcomes across iterations, maximizing experience transfer and continuous improvement.

\paragraph{Strengthen the falsification procedure.}
Our research underscores the importance of falsification to enhance the scientific rigor of the research findings.
While we have prototyped the falsification process in our \AIGSSys system, more efforts are required to strengthen the modules related to knowledge falsification, including the exploitation of the patterns and relationships derived from historical experiments for the guidance of refined research proposals.
Besides, it is also vital for \AIGS systems to investigate whether the delivered new scientific knowledge could generalize across diverse research domains in an autonomous manner.

\paragraph{Expand channels for scientific knowledge dissemination.}
Facilitating the exchange of \AIGSFull is critical, both between humans and AI and among AI systems.
While \citet{lu2024aiscientist} focus on disseminating knowledge through research papers, alternative formats like posters, podcasts, and videos are gaining traction with the rise of multi-modal agents.
Future research should also explore more efficient communication channels between AI systems, beyond structured text or natural language~\citep{pham2023let,chen2024beyond}.

\paragraph{Exploring communication dynamics among autonomous AI researchers.}
As discussed in Section~\ref{sec:literaturereview}, the advancement of AI-accelerated scientific discovery spans four paradigms, culminating in the emergence of an autonomous AI research community (Paradigm IV).
Within this community, individual agentic researchers engage in interactions~\citep{yang2024react} that parallel collaborative dynamics found in human scientific networks.
Analyzing these communication dynamics is essential to understand how fully-autonomous AI agents might effectively collaborate, exchange knowledge, and drive collective progress.
In particular, a deeper exploration of these interactions in a multi-agent system will help establish communication frameworks that support optimal collaboration~\citep{liu2024a}, fostering a robust and productive AI-accelerated research community.

\paragraph{Promote interdisciplinary knowledge integration and experimentation.}
In this work, we primarily focused on the application of \AIGS systems within the domain of machine learning, where experiments could be executed in computers. 
However, future developments should extend these systems to address challenges in other scientific fields, such as biology, which has been preliminarily explored in a concurrent work~\citep{Swanson2024.11.11.623004}, chemistry, and physics, where cross-disciplinary knowledge integration is often crucial.
One major challenge lies in how AI agents can synthesize and align domain-specific knowledge from multiple fields, which often have distinct terminologies, methodologies, and epistemological assumptions. Another critical challenge is the experiment environment, which could be hardly automated and might be highly resource-consuming.  
We hope the integrity and development of optional modules like \textsc{Domain-specific ExpAgent} and \textsc{EnvironmentAgent} mentioned in Section~\ref{sec:overall-system} could alleviate the challenges, and further effort is needed and will be made in future work. 

\section{Ethics and Impact Statement}
\label{sec:ethicandimpact}

In our \AIGSSys system, the agent did not perform harmful operations on computer systems or environment because of the design of \DSL, task constraints and no access to external tools. 
However, while the system developed in this study is limited in scope, \AIGS systems as a whole may have significant impacts in the future, with potential risks that should not be overlooked.
This section explores the potential negative impacts of such systems, drawing on prior research, and offers suggestions for promoting their positive development.

\subsection{Potential Negative Impacts of \AIGS Systems}


\paragraph{Impact on Human Researchers and Academic Community}
In the absence of robust publication standards and academic review processes, \AIGS systems could flood the academic community with low-quality literature, which will further increase researchers' workload and disrupt the efficient dissemination of knowledge~\citep{lu2024aiscientist, si2024can, hu2024nova}.
And although \citet{si2024can} and \citet{kumar2024can} suggest that LLMs can generate ideas more creative than humans, the extent of such creativity remains uncertain.
LLM-powered \AIGS systems tend to rely heavily on existing data and patterns, which could foster \textit{path dependency} and limit opportunities for groundbreaking discoveries.
Additionally, these systems might inadvertently use proprietary or copyrighted material, raising concerns about intellectual property infringement~\citep{kumar2024can}.
Furthermore, \AIGS systems also present several unpredictable challenges for human researchers:
\begin{itemize}
    \item \textbf{Dependence Effect and Cognitive Inertia}: Over-reliance on AI-generated insights may diminish researchers’ independent thinking, leading to cognitive stagnation and a decline in critical thinking skills~\citep{si2024can, hu2024nova}.
    \item \textbf{Ambiguity in Responsibility Attribution}: The involvement of AI complicates the assignment of credit and responsibility, potentially disrupting existing incentive structure~\citep{si2024can, hu2024nova}.
    \item \textbf{Weakened Collaboration and Increased Isolation}: As \AIGS systems become capable of independently generating publishable work, researchers may increasingly rely on these systems, reducing the need for direct collaboration and communication with colleagues. This shift could lead to a decline in interpersonal interaction, weakening traditional research networks built on teamwork and shared discourse~\citep{si2024can, hu2024nova}. Over time, the diminishing frequency of collaborative exchanges may foster a sense of professional isolation among human researchers, heightening the risk of loneliness, disengagement, and reduced psychological well-being.
    \item \textbf{Exacerbated Technological Barriers}: Without equitable access to advanced \AIGS systems, a technological divide could emerge, disadvantaging researchers unfamiliar with or lacking access to these systems, thereby exacerbating inequalities within the community.
\end{itemize}

\paragraph{Impact on Environment}
\AIGS systems can conduct large-scale experiments in parallel, but their dependence on iterative processes carries the risk of inefficient feedback loops, potentially leading to issues such as infinite loops.
This inefficiency, caused by limited reasoning capabilities, the misuse of erroneous information, or ambiguity in task definition~\citep{yang2024towards}, could drive up energy consumption.
Moreover, poorly regulated experiments, especially without adequate simulation environments, can lead to unintended environmental harm.
For example, untested chemical processes in materials science may yield hazardous by-products, while unchecked experiments in nuclear research could increase the risk of radiation leaks~\citep{tang2024prioritizing}.

\paragraph{Impact on Social Security}
\AIGS systems, particularly when compromised by jailbreak attacks, could generate responses that conflict with human values, such as providing instructions for creating explosives.
This raises concerns about their misuse for harmful purposes, such as designing more advanced adversarial attack strategies~\citep{tang2024prioritizing, si2024can, lu2024aiscientist, kumar2024can, hu2024nova}.
Even with benign intentions, unsupervised scientific research may introduce unforeseen societal risks.
For instance, monopolizing breakthroughs in autonomous AI could lead to severe unemployment, market monopolies, and social unrest~\citep{tang2024prioritizing}.

\subsection{Strategies for Responsible and Ethical Development of Automated Research Systems}


\paragraph{Strengthening the Security of Foundation Models}
The most fundamental step in mitigating security risks associated with \AIGS systems is enhancing the security of their foundation models.
Incorporating instructions for handling unsafe research into the alignment training corpus, alongside conducting rigorous safety audits prior to model deployment, are both crucial strategies to ensure the systems be robust and secure~\citep{tang2024prioritizing}.

\paragraph{Aligning Scientific Agents with Human Intentions, Environment and Self-constraints}
Scientific agents in \AIGS systems should align with human intentions, environmental dynamics, as well as self-constraints~\citep{yang2024towards}.
\begin{itemize}
    \item \textbf{Human Intentions}: Agents must accurately interpret user intent, going beyond literal language to capture the deeper purpose of scientific inquiries.
    \item \textbf{Environment}: Agents need to adapt to the specific environments in which they function by applying domain-specific knowledge accurately and utilizing specialized tools effectively.
    \item \textbf{Self-Constraints}: Agents must evaluate task feasibility, manage resources wisely, and minimize waste to ensure sustainable operation. This includes setting boundaries to prevent redundant work or harmful behavior, which is essential for maintaining system efficiency.
\end{itemize}

\paragraph{Providing Comprehensive Training for Human Users}
Comprehensive and rigorous training is essential for users to fully leverage \AIGS systems and prevent unintended consequences~\citep{aisdpi}.
Proper training minimizes the risk of misuse that could lead to environmental harm, resource waste, or unethical research outcomes.
Training programs should focus not only on technical skills but also on ethical considerations, ensuring users understand the limitations and responsibilities associated with these systems~\citep{tang2024prioritizing}.

\paragraph{Building a Collaborative Framework Between Automated Research Systems and Human Researchers}
To prevent \AIGS systems from exerting excessive influence on the academic community, collaboration between \AIGS systems and human researchers will play a crucial role~\citep{si2024can, hu2024nova}.
It is essential to explore the new roles and responsibilities that human scientists may need to assume in this evolving research landscape shaped by the presence of \AIGS systems.
A well-structured partnership can leverage the complementary strengths of both, enabling outcomes that neither could achieve independently.
Moreover, such collaboration fosters interaction among human researchers, encouraging deeper communication and mitigating the sense of isolation that may arise from increased reliance on automated tools.

\paragraph{Establishing Comprehensive Legal and Accountability Frameworks}
A robust legal and accountability framework is crucial to govern the use of \AIGS systems. This framework should:
\begin{itemize}
    \item \textbf{Define Clear Scientific Research Boundaries}: Specify the permissible scope and limitations of these systems, where regulate agents with the \DSL might be helpful.
    \item \textbf{Clarify Responsibility and Credit Allocation}: Establish guidelines for assigning credit and responsibility for research outcomes generated with the assistance of \AIGS systems~\citep{si2024can, hu2024nova}.
    \item \textbf{Implement Penalties for Misuse}: Outline liability measures and penalties to address harmful behavior or unethical practices involving these systems.
\end{itemize}

\paragraph{Using \AIGS Systems to Address Its Own Challenges}
\AIGS systems can also play a proactive role in addressing the challenges and even ethical issues introduced by themselves. 
For example, \AIGS systems could be used to monitor and evaluate outputs from other automated systems, identifying potential ethical issues, biases, or environmental risks before they escalate.
Moreover, \AIGS systems can facilitate the development of guidelines, by automating the analysis of research trends and regulatory needs, thus helping shape future policies for responsible AI use.
When employed strategically, \AIGS systems become not only tools for discovery but also mechanisms for self-regulation, creating a virtuous cycle of innovation and governance.

\section{Conclusion}

We introduce the concept of \textbf{\AIGS} in this paper and implement \AIGSSys, a baby-step toward full-process automated scientific discovery systems, with a focus on incorporating \textit{\textbf{falsification}} into the research process. By integrating a \FalsificationAgent, the multi-agent system can identify and verify potential discoveries. Techniques as \DSL and multi-sampling strategy are introduced for two other principles of \AIGS systems design, \textit{\textbf{executability}} and \textit{\textbf{creativity}}. Preliminary experiments show promise, though the system's performance remains below that of experienced human researchers. This work lays the groundwork for future developments in \AIGS systems, with further improvements over \AIGSSys and ethical considerations necessary for advancing the field.

\bibliography{iclr2025_conference}
\bibliographystyle{iclr2025_conference}

\pagebreak

\appendix
\section{Implementation Details of the \AIGSSys system}
\label{app:sys-detail}

In this section, we elaborate the implementation details of the \AIGSSys system. All artifacts are used as intended with their license strictly followed in our work.

\subsection{Research-Agnostic Implementation}

\paragraph{System Pipeline}
We posit that all agents mentioned in Section~\ref{sec:overall-system} contribute to a full-process \AIGS system, but based on preliminary experiments, we simplify the design of \ExperimentAgent and \LiteratureAgent to a large extent in our implementation. For \ExperimentAgent, given the design of \DSL with human effort, proposed methodology generated by \ProposalAgent can be executed reliably in experiments, which is also shown in Section~\ref{tab:success_rate}. This reduces the need of iteratively refining proposals between \ProposalAgent and \ExperimentAgent. For \LiteratureAgent, preliminary results show literature integration did not significantly impact the outcomes in both phases of \AIGSSys. We conclude the reason as that agents failed to understand the in-depth literature information and the retrieval of literature did not match the need of each agent perfectly. Therefore, in our implementation, we minimize the design of these two agents: \ExperimentAgent functions through fixed code, and \LiteratureAgent was not put into pratical use. Other optional agents are designed to function in broader research fields, and we chose to omit them in experiments based on the selected research topics for experiments (Section~\ref{sec:exp-setup}).

\paragraph{Hyper-Parameters}
Experiments in ICL (In-Context Learning) of the data engineering research and in language modeling research are conducted on 8 NVIDIA GeForce RTX 3090 24 GB GPUs. Experiments in SFT (Supervised Fine-tuning) of the data engineering research and in Self-Instruct alignment research are conducted on 8 A100 80GB GPUs. All researches utilize the gpt-4o-2024-05-13 model as the underlying model for our agents. When agents invoke GPT-4o, we use the openai module\footnote{\url{https://github.com/openai/openai-python}} with a temperature setting of 0.7, while all other parameters are setting as default values. During the synthesis of proposals, \ProposalAgent generates three sets of proposals with a temperature of 0.7. After generation, the Jaccard similarity~\citep{jaccard1901etude} of bigram sets is calculated between the methodology of each proposal and the methodology produced in the previous iteration. The proposal with the lowest similarity in methodology is selected as the final output to increase its diversity. For \ReviewAgent and \FalsificationAgent, they invoke the GPT-4o only once each time when generating responses.

\subsection{Research-Specific Implementation}

\paragraph{Data Engineering}
In this research experiment, our system is tasked with exploring different approaches to improve the quality of Alpaca-GPT4 dataset~\citep{peng2023instruction}. The \DSL configuration and instance are shown in Figure~\ref{fig:dsl} and Figure~\ref{fig:dsl-1}. The Llama-3-8B-Instruct\footnote{\url{https://huggingface.co/meta-llama/Meta-Llama-3-8B-Instruct}} model is employed to rate all data samples with the principles in \DSL. We deploy Llama-3-8B-Instruct using vLLM\footnote{\url{https://github.com/vllm-project/vllm}.}, configuring the temperature to 0.05, while keeping all other parameters at the default settings. We use Llama-3-8B\footnote{\url{https://huggingface.co/meta-llama/Meta-Llama-3-8B}.} for ICL- and SFT-alignment, and the model and the fine-tuned checkpoints are deployed using vLLM with a maximum token limit of 1024, while other parameters follow the default configurations provided by FastChat\footnote{\url{https://github.com/lm-sys/FastChat}.}. 
In falsification process, the \AIGSSys system identifies the factors that contribute to quality improvements and conclude whether there are ways to stably improve the quality of the extracted dataset, thus delivering valuable scientific discoveries. 
For significance screening in \FalsificationAgent, iterations are identified as having significant improvements if the difference of adjacent benchmarking results exceeds 1.5 for the ICL-aligned Llama-3-8B on the Vicuna-Bench (the validation benchmark) or 0.5 on the MT-Bench (the test benchmark). From these iterations, candidates for scientific discovery are extracted. For hyper-parameters, we set the total iteration number $M=5$ and set the multi-sample threads number $N=32$.

\paragraph{Self-Instruct Alignment}
In this research experiment, our system is tasked with exploring different approaches to improve the quality of synthesized SFT data from a seed dataset in Self-Instruct\footnote{\url{https://github.com/yizhongw/self-instruct}.}~\citep{wang-etal-2023-self-instruct}. We use GPT-4o to rewrite the seed data for better quality with the temperature parameter set to 0.05. The \DSL configuration and instance are shown in Figure~\ref{fig:dsl} and Figure~\ref{fig:dsl-2}. We use the Llama-3-8B\footnote{\url{https://huggingface.co/meta-llama/Meta-Llama-3-8B}.} model to generate instructions and responses, with it also serving as the base model for SFT alignment. We use LoRA~\citep{hu2022lora} method from LLaMA-Factory\footnote{\url{https://github.com/hiyouga/LLaMA-Factory}.} to fine-tune the model with default training hyper-parameters\footnote{\url{https://github.com/hiyouga/LLaMA-Factory/blob/main/examples/train_lora/llama3_lora_sft.yaml}.}. The other experiment setting is the same as data engineering research. For hyper-parameters, we set the total iteration number $M=15$ and set the multi-sample threads number $N=1$ due to limited computing resources for parallel model training.

\paragraph{Language Modeling}
In this research, the system is tasked to pre-train a mini-sized language model on several small corpora, aiming to improve performance by minimizing loss on the selected datasets. The experiment mainly follows the same setup as the language modeling task in AI Scientist~\citep{lu2024aiscientist}, based on the nanoGPT project~\footnote{\url{https://github.com/karpathy/nanoGPT}.}. The \DSL configuration and instance are shown in Figure~\ref{fig:dsl} and Figure~\ref{fig:dsl-3}, where we guide the models in adjusting parameters related to model architecture and training process. For the experiments, we use the sampling scripts provided in the template code without modifications. For hyper-parameters, we set the total iteration number $M=10$ and set the multi-sample threads number $N=1$ due to limited computing resources for parallel model training.

\section{Experiment Details}\label{app:exp-detail}

\subsection{Guidelines for Human Evaluators}
\label{sec:human_evaluator}

To thoroughly assess the quality of our falsification process, we conducted a human evaluation of 20 agent-generated falsification logs. The guidelines are summarized as follows:

\begin{itemize}
    \item \textit{\textbf{Importance Score}}: Assess the significance of the proposed scientific discovery candidate, considering its potential impact on experimental results and its relevance and consistency with the main experiments.
    \item \textit{\textbf{Consistency Score}}: Evaluate whether the proposed ablation experiments align with the scientific discovery candidate and whether the experiment appropriately isolates the factor in question.
    \item \textit{\textbf{Correctness Score}}: Determine whether the final scientific discovery drawn from the falsification process is correct based on the ablation and baseline results.
\end{itemize}

For each dimension, the evaluator assigns an integer score ranging from 0 to 2, where a higher score indicates better performance. The overall statistic results are shown in Table~\ref{tab:human}. 

\subsection{API Costs of the Full-Process Research Experiment}\label{app:token-anal}
In our experiments, we measured the average token counts and costs of different phases of \AIGSSys (Section~\ref{sec:overall-system}) for invoking the GPT-4o API and the results are presented in Table~\ref{table:costs}. Note that as the experimental records in past iterations are used as input in most requests, with the rise of iteration, the length of record will consequently increase, leading to the use of more tokens. 

\begin{table}[h!]
\centering
\begin{tabular}{lccc}
\toprule
                    & \textbf{Input Tokens}  & \textbf{Generated Tokens} & \textbf{Cost} (\$) \\ \midrule\midrule
\textbf{Pre-Falsification} (per iter.)               & 6,616.2              & 761.5                & 0.045       \\ 
\textbf{Falsification} (per disc. cand.)               &  43,375.5          &  1,120.3             & 0.234     \\ 
\bottomrule
\end{tabular}
\caption{Average token consumption and API costs for GPT-4o API in the full-process research experiment. The costs at pre-falsification phase is calculated for each iteration, and the costs at falsification phase is calculated for each discovery candidate.} 
\label{table:costs}
\end{table}

\subsection{\DSL Demonstrations for Different Research Topics}
\begin{figure}[htbp]
    \centering\includegraphics[width=\linewidth]{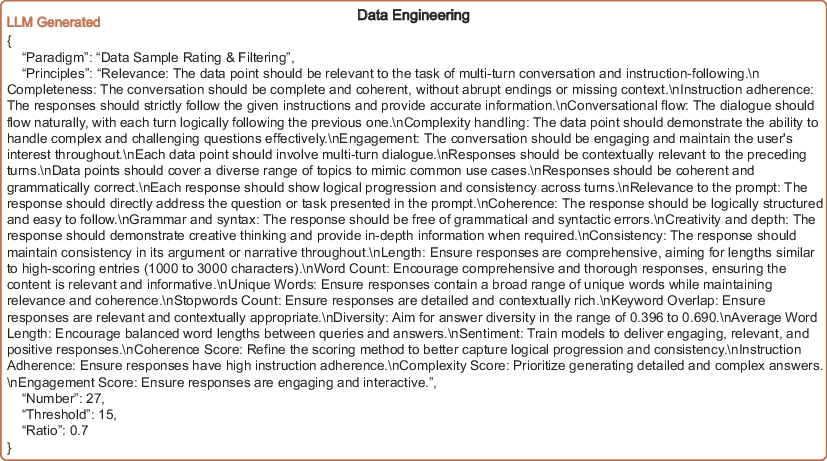}
    \caption{The \DSL instance for data engineering research.} 
    \label{fig:dsl-1}
\end{figure}

\begin{figure}[htbp]
    \centering\includegraphics[width=\linewidth]{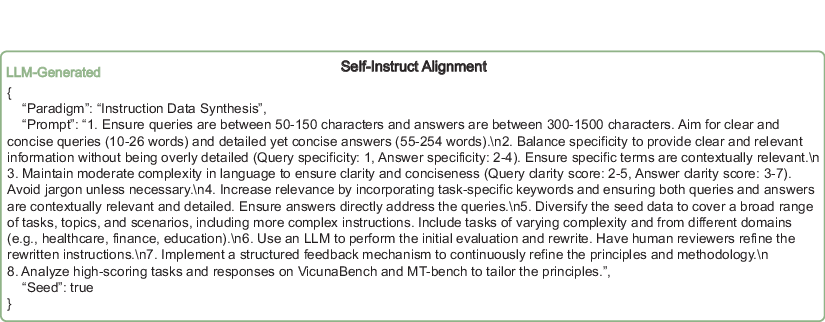}
    \caption{The \DSL instance for self-instruct alignment research.} 
    \label{fig:dsl-2}
\end{figure}

\begin{figure}[htbp]
    \centering\includegraphics[width=0.4\linewidth]{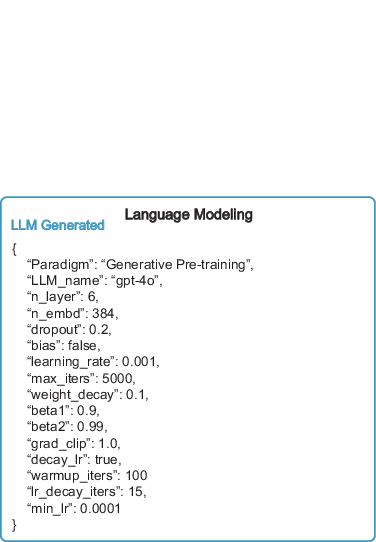}
    \caption{The \DSL instance for language modeling research.} 
    \label{fig:dsl-3}
\end{figure}

We present an example of the methodology in \DSL format generated during the experiment for each research topic, as shown in Figure~\ref{fig:dsl-1}, Figure~\ref{fig:dsl-2} and Figure~\ref{fig:dsl-3}, corresponding to data engineering, self-instruct alignment, and language modeling, respectively. 

\section{Prompting Structure}\label{app:prompt}

\begin{figure}[htbp]
    \centering\includegraphics[width=0.9\linewidth]{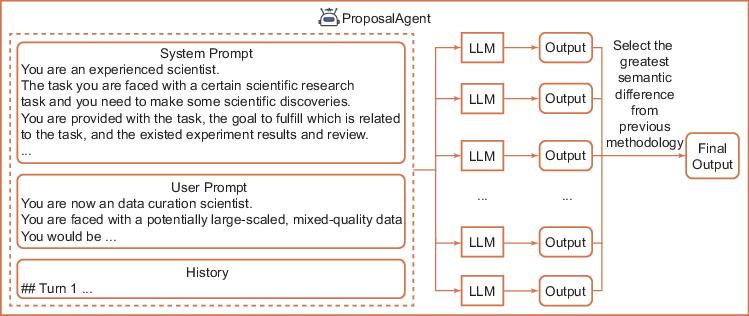}
    \caption{The prompting structure for the \ProposalAgent includes a general system prompt, a research-topic-specific user prompt and history logs. The LLM generates multiple outputs, covering elements such as idea, methodology, \DSL, etc. From these outputs, the one whose methodology has the greatest semantic difference from the previous round’s methodology is selected as the idea for the current round, aiming to boost creativity in ideation.}
    \label{fig:prompt-proposal}
\end{figure}

\begin{figure}[htbp]
    \centering\includegraphics[width=\linewidth]{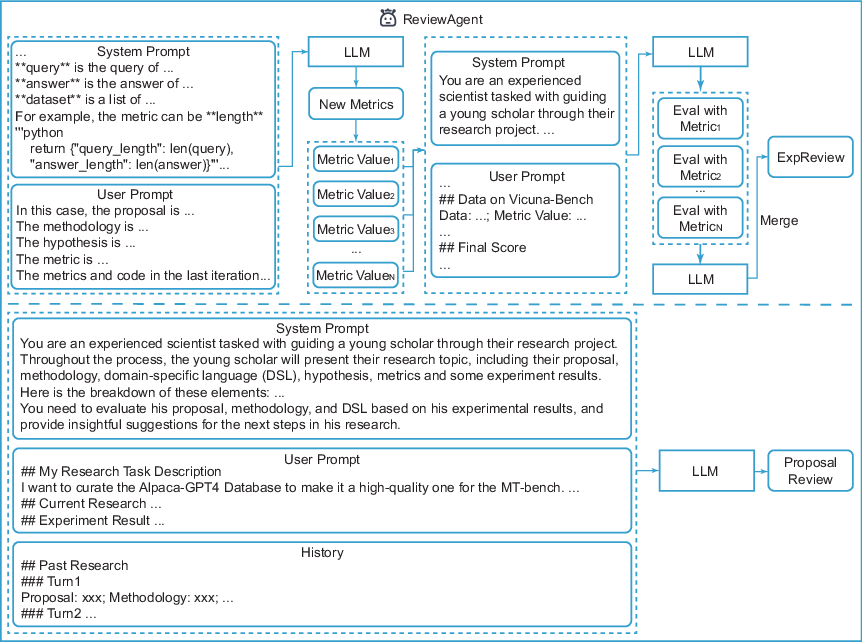}
    \caption{The \ReviewAgent will first generate new metrics and then analyze each metric individually using the LLM. Following this, the \ReviewAgent will call the LLM to merge the analysis results for each metric, resulting in the \textit{ExpReview}. Next, the \ReviewAgent will assess the experimental results by integrating insights from previous ideas and experiments, yielding the \textit{ProposalReview}.}
    \label{fig:prompt-review}
\end{figure}

\begin{figure}[htbp]
    \centering\includegraphics[width=\linewidth]{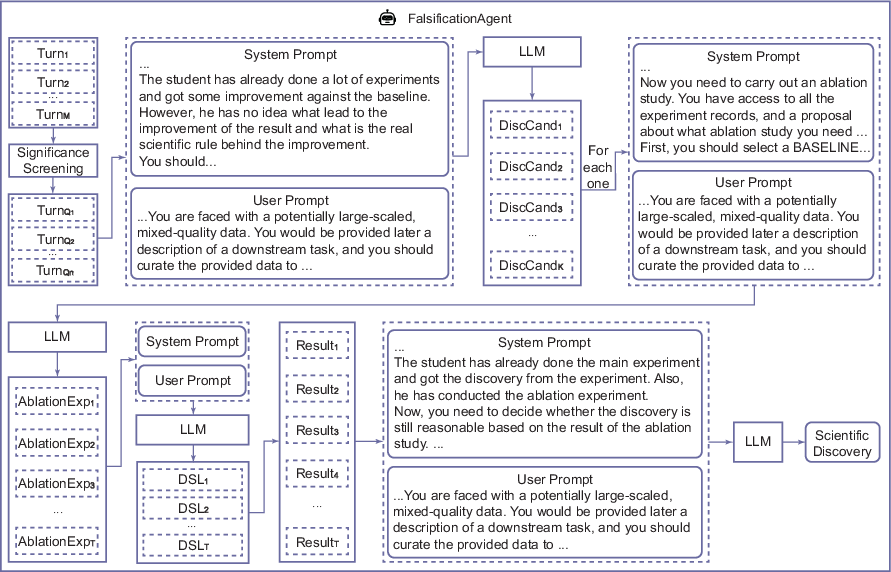}
    \caption{The \FalsificationAgent first screens all history turns to identify turns with notable changes in results. It then generates discovery candidates from the results obtained through significance screening. For each discovery candidate, it then creates several ablation experiment setups and generates the corresponding \DSL to obtain experimental results. Once the experimental results are obtained, the \FalsificationAgent calls on the LLM to produce the final scientific discovery.}
    \label{fig:prompt-falsification}
\end{figure}

In this section, we will briefly introduce the prompting structures of the \ProposalAgent, \ReviewAgent, and \FalsificationAgent as shown in Figure~\ref{fig:prompt-proposal}, Figure~\ref{fig:prompt-review}, and Figure~\ref{fig:prompt-falsification}, respectively. For detailed prompts, please refer to our code repository\footnote{\url{https://github.com/AgentForceTeamOfficial/Baby-AIGS}.} .

\end{document}